\documentclass[journal]{IEEEtran}
\usepackage{amsmath,amssymb,amsfonts}
\usepackage{cases}
\usepackage{color,xcolor}
\usepackage{graphicx}
\usepackage{subfigure}
\usepackage{algorithm}
\usepackage{makecell}
\usepackage{algorithmic}
\usepackage{epstopdf}
\usepackage{multirow}
\usepackage{rotating}
\usepackage{booktabs}
\usepackage{tikz}
\usepackage{pdflscape}
\usepackage{cite}
\usepackage{enumitem} 
\usepackage{bm}
\usepackage{newtxmath}
\usepackage[colorlinks,
linkcolor=blue,
anchorcolor=blue,
citecolor=blue
]{hyperref}

\newtheorem{Theorem}{Theorem}[section]

\newtheorem{Remark}[Theorem]{Remark}

\graphicspath{{figures/}}

\begin{document}

\title{Adaptive Multi-Order Graph Regularized NMF with Dual Sparsity for Hyperspectral Unmixing}

\author{Hui Chen, \IEEEmembership{Member,~IEEE}, Liangyu Liu, Xianchao Xiu, and Wanquan Liu, \IEEEmembership{Senior Member,~IEEE}

\thanks{This work was supported in part by the Natural Science Foundation of Shanghai under Grant 24ZR1425700, the National Natural Science Foundation of China under Grant 12371306, and in part by the Project of the State Administration of Foreign Experts under Grant H20240974. (\textit{Corresponding author: Xianchao Xiu}.)}
\thanks{H. Chen and L. Liu are with the School of Automation Engineering, Shanghai University of Electric Power, Shanghai 200090, China (e-mail: chenhui@shiep.edu.cn; liangyuliu@mail.shiep.edu.cn).}
\thanks{X. Xiu is with the School of Mechatronic Engineering and Automation,  Shanghai University,  Shanghai 200444,  China (e-mail: xcxiu@shu.edu.cn).}
\thanks{W. Liu is with the School of Intelligent Systems Engineering, Sun Yat-sen University, Guangzhou 510275, China (e-mail: liuwq63@mail.sysu.edu.cn).}
}

\maketitle
\begin{abstract}
Hyperspectral unmixing (HU) is a critical yet challenging task in remote sensing.
However, existing nonnegative matrix factorization (NMF) methods with graph learning mostly focus on first-order or second-order nearest neighbor relationships and usually require manual parameter tuning, which fails to characterize intrinsic data structures. 
To address the above issues, we propose a novel adaptive multi-order graph regularized NMF method (MOGNMF) with three key features.  
First, multi-order graph regularization is introduced into the NMF framework to exploit global and local information comprehensively.
Second, these parameters associated with the multi-order graph are learned adaptively through a data-driven approach.
Third, dual sparsity is embedded to obtain better robustness, i.e., $\ell_{1/2}$-norm on the abundance matrix and $\ell_{2,1}$-norm on the noise matrix.
To solve the proposed model, we develop an alternating minimization algorithm whose subproblems have explicit solutions, thus ensuring effectiveness. 
Experiments on simulated and real hyperspectral data indicate that the proposed method delivers better unmixing results.
\end{abstract}


\section{Introduction}

\IEEEPARstart{I}{n} the field of remote sensing, hyperspectral images (HSIs) have attracted extensive attention due to their high spatial and spectral resolution. However, HSIs are usually high-dimensional and noisy, which has led to a series of techniques to extract useful information, such as denoising \cite{dong2025hyperspectral,chen2024fast,zhang2023hyperspectral}, classification \cite{nie2024structure,li2024mambahsi,datta2022hyperspectral}, anomaly detection  \cite{zhang2024light,xiao2024hyperspectral,liu2025exploiting}, and hyperspectral unmixing (HU)  \cite{rasti2024image,li2024sparse,gao2025ssaf}. Among them, HU plays a critical role in analyzing HSIs by separating spectral endmembers and estimating their abundance fractions, with wide applications in medical diagnostics \cite{ul2021review}, environmental monitoring \cite{esi2024nonnegative}, and precision agriculture \cite{ram2024systematic}.

Based on the spectral mixing mechanism, a large number of methods have been proposed to deal with the HU problem. The linear mixing model (LMM) and the nonlinear mixing model (NLMM) are two representatives. LMM assumes that each pixel is a linear combination of endmember spectra, while NLMM further accounts for multiple scattering and microscopic mixing effects \cite{6816071}. Typical NLMM methods include the generalized bilinear model  \cite{halimi2011nonlinear}, the p-order polynomial model \cite{marinoni2015novel}, and the multilinear mixing model \cite{heylen2015multilinear}. Due to the inherently nonconvex nature of unmixing problems, nonlinear methods often involve high degrees of freedom, which may lead to the generation of physically meaningless endmembers \cite{10870282}. Nowadays, LMM has been widely regarded as a reasonable first-order approximation of the physical measurement process due to its strong physical interpretability\cite{dobigeon2013nonlinear}.

Generally speaking, LMM can be divided into geometry-based \cite{li2015minimum,li2024initially}, statistics-based \cite{wang2014abundance,lu2019subspace}, regression-based \cite{ince2020superpixel,deng2023robust}, and deep learning-based \cite{deshpande2021practical,chen2023improved}. Geometry-based methods like vertex component analysis (VCA) \cite{nascimento2005vertex} extract endmembers by locating extreme pixels according to the data’s simplex structure, but they usually require pure pixels. Regression-based methods such as fully constrained least squares (FCLS) \cite{heinz2001fully} estimate abundances under nonnegativity and sum-to-one constraints. The combination of VCA and FCLS, known as the VCA-FCLS framework, is widely used as initialization by separately handling endmember extraction and abundance estimation. In contrast, nonnegative matrix factorization (NMF) \cite{lee1999learning} jointly estimates endmembers and abundances under nonnegativity constraints, aligning with the physical properties of hyperspectral data. Unlike VCA-FCLS, it does not require pure pixels and supports unsupervised learning. Its flexibility and extensibility have made NMF a widely used tool, as highlighted in recent surveys \cite{gillis2020nonnegative, feng2022hyperspectral}.
 
\begin{figure*}[t]  
    \centering
    \includegraphics[width=1\textwidth]{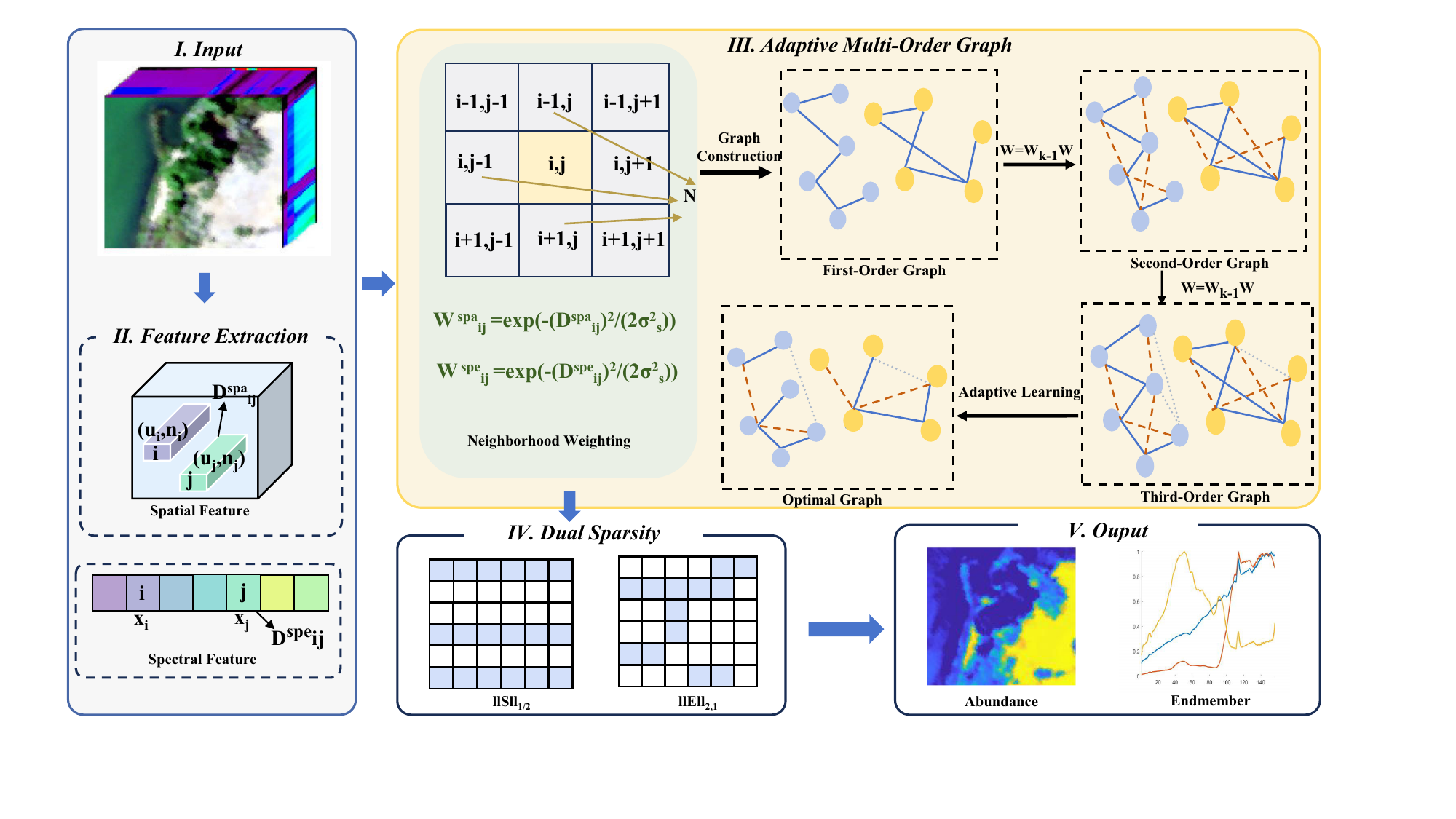}
    \vspace{-1.7cm}
     \caption{Flowchart of our proposed method. First, input the HSI. Second, extract spatial and spectral features. Third, construct the multi-order graph, where blue solid lines, yellow dashed lines, and gray dashed lines represent first-order neighbor relationships,  second-order neighbor relationships, and  third-order neighbor relationships, respectively. Fourth, add dual sparse regularization terms. Finally,  output the abundance map and the endmember map.}
    \label{Flowchart}
\end{figure*}

To enhance the unmixing performance of NMF, researchers have developed a large number of constrained NMF methods and achieved remarkable results. For example,  Miao and Qi \cite{miao2007endmember} incorporated a volume constraint based on the convex cone simplex into the standard NMF and proposed MVC-NMF. Qian et al. \cite{qian2011hyperspectral} suggested $\ell_{1/2}$-norm \cite{xu2012l_} constrained NMF to improve the interpretability, which is denoted as SNMF in our paper without ambiguity. Subsequently, Zhu et al. \cite{zhu2014spectral} replaced $\ell_{1/2}$-norm with $\ell_q$-norm $(0<q<1)$ to obtain the ability of adaptive selection \cite{liu2024efficient}. Furthermore, Wang et al. \cite{wang2017spatial} considered the sparse structure of local pixels and combined it with the smoothness constraint, which is called spatial group sparsity regularized NMF (SGSNMF). Besides, there are low-rank constrained NMF \cite{ye2022combining}, \cite{xu2024spectral}, abundance separation constrained NMF \cite{liu2010approach}, \cite{liu2016hyperspectral}, etc. We would like to point out that although constrained NMF methods significantly improve the performance of traditional NMF by incorporating prior information, it is still not robust to noise and outliers \cite{li2017robust,peng2021robust}. To this end, Févotte et al. \cite{fevotte2015nonlinear} proposed robust NMF (RNMF) by introducing additional sparse matrices to filter out sparsely distributed noise. Recently, Zhang et al. \cite{zhang2024matrix} constructed group sparse constrained NMF, which is denoted as IRNMF. In addition to the above constrained NMF methods, Abdolali et al. \cite{abdolali2024dual} transformed the minimum-volume NMF problem in the primal space to a maximum-volume problem in the dual space for a more stable global solution, which is called MV-DUAL. 
In summary, the above-mentioned sparse constrained NMF methods have good unmixing performance, hence we will continue to align this strategy.

Recently, graph learning was incorporated with  NMF to capture the low-dimensional features in HSIs \cite{duan2020local,guo2021sparse}. Lu et al. \cite{lu2012manifold} proposed a graph-regularized NMF method by combining manifold regularization with sparse constrained NMF, where the manifold regularization term was validated to strengthen the connection between original images and abundance maps. Yang et al. \cite{yang2015geometric} considered spatial geometric distance and spectral geometric distance as metrics and obtained a space-spectral manifold-regularized NMF model. Wang et al. \cite{wang2016hypergraph} introduced the hypergraph structure to capture the similarity relationships among spatially neighboring pixels and proposed hypergraph-regularized sparse NMF. Very recently, Qu et al. \cite{qu2023hyperspectral} presented a high-order graph regularized NMF method,  called HGNMF-FS. However, these graph learning-based methods have two limitations. On the one hand, most methods rely on first-order or fixed-order neighborhoods, resulting in sparse graphs that fail to capture high-order pixel associations \cite{meilua2024manifold}. This limitation reduces the model's ability to represent both global and local structures, such as the similarity between neighboring pixels and the consistency across distant but similar regions \cite{tang2015line}. On the other hand, their graph construction typically depends on predefined parameters, lacking the flexibility to adaptively learn graph structures.  As shown in Fig. \ref{Flowchart}, adaptive learning allows the model to discover the optimal graph as a graph regularization constraint, where the complementary nature of different orders significantly enhances the expressiveness of the graph representation.

In addition, deep learning can effectively model complex nonlinear mixtures, thereby capturing hierarchies in hyperspectral data. Tao et al. \cite{tao2024abundance} addressed the strong reliance of existing unmixing methods on initialization, and proposed an abundance-guided attention network called A2SAN. Chen et al. \cite{chen2025unmamba} leveraged the Mamba model to achieve efficient hyperspectral unmixing with fewer parameters and lower computational cost. Very recently, Che et al. \cite{che2024robust} learned multi-order graphs from different views, thereby capturing high-order relationships between data. Yang et al. \cite{yang2024auto} embedded graph regularization constraints in each view and adaptively assigned weights to different views. These two methods can effectively extract valuable information from multi-view data.

Motivated by the studies mentioned above, we propose a novel robust and adaptive multi-order graph regularized NMF method for HU. As illustrated in Fig. \ref{Flowchart}, this method can adaptively learn spatial and spectral graph structures according to the characteristics of the input HSIs, thereby efficiently capturing data features. In addition, dual sparse regularization terms are incorporated to enhance  reliability, i.e., $\ell_{1/2}$-norm on the abundance matrix and $\ell_{2,1}$-norm on the noise matrix. 
In summary, the main contributions of this article are as follows.

\begin{enumerate}
\item We introduce multi-order graphs to capture the similarity between pixels by defining spatial and spectral geometric distance, thereby solving the problem that existing unmixing methods only rely on single-order nearest neighbor relations and cannot fully capture the complex relationship between pixels in HSIs.
\item We design an adaptive strategy to learn the weights of multi-order graphs and leverage regularization techniques to avoid overfitting. On this basis, we further incorporate dual sparsity into the NMF framework and finally establish a novel HU method.
\item We develop an efficient alternating minimization algorithm whose subproblems  can be easily calculated by fast solvers. Extensive experiments on simulated and real HSIs verify the effectiveness of the proposed method and the necessity of multi-order graphs.
\end{enumerate}

The remainder of this article is organized as follows. 
Section \ref{sec:notations} introduces the notations and related works. 
Section \ref{sec:Multi-Order Graph} describes the proposed model in detail.
Section \ref{Optimization Algorithm} outlines its optimization algorithm. 
Section \ref{Numerical Experiments} provides the experimental results.
Section \ref{sec:CONCLUSION} concludes this article.

\section{Preliminaries }\label{sec:notations}
\subsection{Notations}
In this paper, matrices are denoted by bold capital letters, vectors by boldface characters, and scalars by lowercase letters. Let \( \mathbb{R}^{D} \) and \( \mathbb{R}^{M\times N} \) refer to the sets of all \( D \)-dimensional vectors and \( M \times N \)-dimensional matrices, respectively. The \( i \)-th element of a vector \( \mathbf{x} \in \mathbb{R}^{D} \) is written as \( x_i \). 
For a matrix \( \mathbf{X} \in \mathbb{R}^{M \times N} \), \( X_{ij} \) denotes its \( ij \)-th element, with \( \mathbf{x}^i \) and \( \mathbf{x}_j \) indicating the \( i \)-th row and the \( j\)-th column, respectively.  
The transpose of \( \mathbf{X} \) is written as \( \mathbf{X}^\top \) and the trace is \( \mathrm{Tr}(\mathbf{X}) \).
In addition, the Frobenius norm is denoted by \( \|\mathbf{X}\|_F  \), and the \( \ell_{1/2} \)-norm and  \( \ell_{2,1} \)-norm  are, respectively, defined by
\begin{equation}
\| \mathbf{X} \|_{1/2} = \sum_{i=1}^M  \sum_{j=1}^N  |X_{ij}|^{1/2},~\| \mathbf{X} \|_{2,1} = \sum_{i=1}^M \| \mathbf{x}^i \|_2.
\end{equation}

 \subsection{Linear Mixing Model}
Assume that the spectral response of a pixel is a mixture of multiple endmembers, where the proportion of each endmember corresponds to its abundance, the linear mixing model (LMM) can be characterized by 
\begin{equation}
\bm{\mathbf{X}} = \bm{\mathbf{A}} \bm{\mathbf{S}} + \bm{\mathbf{E}},
\end{equation}
where $\mathbf{X}=[\mathbf{x}_1, \ldots, \mathbf{x}_N] \in \mathbb{R}^{L \times N}$ is the spectral matrix that consists of all pixel spectral vectors, with $L$ and $N$ being the number of spectral bands and pixels, respectively. $ \mathbf{A} \in \mathbb{R}^{L \times M}$ is the endmember matrix, where $M$ represents the total number of endmembers. $\mathbf{S} \in \mathbb{R}^{M \times N}$ is the abundance matrix, which describes the proportion of each endmember within each pixel. $\mathbf{E} \in \mathbb{R}^{L \times N}$ is the  noise matrix.

As for LMM, the abundances should satisfy the abundance nonnegativity constraint (ANC) and the abundance sum-to-one constraint (ASC). Specifically, ANC requires that each element in the abundance vector of each pixel must be nonnegative, because the abundances of different materials represent their contribution rates and cannot be negative. ASC requires that the sum of all elements in the abundance vector of each pixel must be one, because the total abundance of each material should completely cover the spectral response of the pixel.

\subsection{Nonnegative Matrix Factorization}
Under the assumption of LMM, nonnegative matrix factorization (NMF) is a popular method for solving HU problems. Given a nonnegative matrix \( \mathbf{X} \in \mathbb{R}^{L \times N} \), NMF aims to decompose it into two nonnegative matrices, i.e., \( \mathbf{A} \in \mathbb{R}^{L \times M}\) and \( \mathbf{S} \in \mathbb{R}^{M \times N} \), where \( M < \min(L, N) \). The mathematical model can be described as

\begin{equation}
      \begin{aligned}
	\min_{\mathbf{A},\mathbf{S}} \quad & \frac{1}{2} \| \mathbf{X} - \mathbf{A}\mathbf{S} \|_F^2 \\
	\rm{s.t.}~~~& \mathbf{A} \geq 0, ~\mathbf{S} \geq 0.
    \end{aligned}
\end{equation}	

In order to deal with the above problem, an effective alternating iterative method is proposed, called the multiplicative update rule (MUR) \cite{lee2000algorithms}, i.e., 
\begin{equation}
	\begin{aligned}
\mathbf{A}& \leftarrow \mathbf{A}.* \left( \mathbf{X} \mathbf{S}^\top \right) ./ \left( \mathbf{A} \mathbf{S} \mathbf{S}^\top \right),\\
\mathbf{S}& \leftarrow \mathbf{S}.* \left( \mathbf{A}^\top \mathbf{X} \right) ./ \left( \mathbf{A}^\top \mathbf{A}  \mathbf{S} \right),
	\end{aligned}
\end{equation}
where the operator $\cdot \ast$ and $./$ represent element-wise multiplication and division, respectively. 

Since it is a nonconvex problem, there may exist some local optimal solutions. As a result, when applying NMF to practical problems, it is necessary to consider the physical meanings of \( \mathbf{A} \) and \( \mathbf{S} \), and add the corresponding constraints in the objective function to align with these meanings.

\subsection{Graph Constrained NMF}

It is well known that graph learning can be used to project high-dimensional data into a low-dimensional subspace while preserving the geometric relationship between data points. HGNMF-FS, as a representative constrained NMF method, integrates high-order graph regularization, adaptive feature selection, and $\ell_{1/2}$-norm regularization into the NMF framework, which can be expressed as
\begin{equation}\label{HGNMF-FS}
	\begin{aligned}
		\min_{\mathbf{A}, \mathbf{S}, \mathbf{\Lambda}} \quad & \frac{1}{2} \| \mathbf{\Lambda} (\mathbf{X} - \mathbf{A S}) \|_F^2 + \gamma\| \mathbf{S} \|_{1/2} + \frac{\lambda}{2} \mathrm{Tr}(\mathbf{S} \mathbf{L}_h \mathbf{S}^\top)\\
        	\rm{s.t.}~~~~& \mathbf{A} \geq 0, ~\mathbf{S}  \geq 0, ~\mathbf{\Lambda} \geq 0, ~\mathbf{1}_{M}^{\top } \mathbf{S}= \mathbf{1}_{N}^{\top}.
	\end{aligned}
\end{equation}
Note that \( \mathbf{L}_h = \beta_1 \mathbf{L}_1 + \beta_2 \mathbf{L}_2 \), where \( \mathbf{L}_1 \) and \( \mathbf{L}_2 \) represent the first-order and second-order Laplacian matrices, respectively, and $\beta_1, \beta_2$ are the parameters. Besides, \( \mathbf{\Lambda} \) can be adaptively adjusted during the unmixing process as stated in \cite{qu2023hyperspectral}.

 \section{The Proposed Model}\label{sec:Multi-Order Graph}
 
This section first describes spatial and spectral graphs, then proposes a novel multi-order graph framework, followed by our proposed model for the HU task.

 \subsection{Spatial and Spectral Graphs}

Generally speaking, spatial graphs are used to measure the spatial similarity between pixels, and spectral graphs are used to quantify the spectral similarity between pixels, thus they can provide complementary information to better capture the global and local structures of HSIs.

Given two pixels $x_i$ and $x_j$ with spatial coordinates $(u_i, n_i)$ and $(u_j, n_j)$, respectively, their Euclidean distance and the corresponding weight matrix are typically derived using the following heat kernel function
\begin{equation}
	\begin{aligned}
D_{ij}^{\text{spa}} & = \sqrt{(u_i - u_j)^2 + (n_i - n_j)^2},\\
W_{ij}^{\text{spa}}  & = \exp\left(-\left(D_{ij}^{\text{spa}}\right)^2 / (2\sigma_s^2)\right),
	\end{aligned}
\end{equation}
where $\sigma_s$ is the spatial scaling parameter that controls the influence of the distance between neighboring pixels. In this way, the spatial graph can capture the local spatial structure of pixels, while Gaussian weighting can smooth the similarity between neighboring pixels.

Considering the spectral vectors $ \mathbf{x}_i $ and $ \mathbf{x}_j $, their Euclidean distance and spectral weight matrices are usually given by
\begin{equation}
	\begin{aligned}
D_{ij}^{\text{spe}} & = \|\mathbf{x}_i - \mathbf{x}_j\|_2, \\
W_{ij}^{\text{spe}}  & =\exp\left(-\left(D_{ij}^{\text{spe}}\right)^2 / (2\sigma_l^2)\right),
	\end{aligned}
\end{equation}
where $\sigma_l$ is the spectral scaling parameter that controls the range of spectral similarity. The Gaussian kernel ensures that pixels with high spectral similarity receive higher weights, while different  pixels have  less influence on unmixing.

Therefore, the corresponding first-order Laplacian matrices of the spatial and spectral graphs are defined as
\begin{equation}
	\begin{aligned}
	\textbf{L}^{\text{spa}} &= \textbf{D}^{\text{spa}} - \textbf{W}^{\text{spa}},\\
\textbf{L}^{\text{spe}} &= \textbf{D}^{\text{spe}} - \textbf{W}^{\text{spe}},
	\end{aligned}
\end{equation}
where \( \mathbf{D}^{\text{spa}} \) and \( \mathbf{D}^{\text{spe}} \) represent diagonal matrices, and their diagonal elements are, respectively, given by
\begin{equation}
 D_{ii}^{\text{spa}} = \sum_{j=1}^{n} W_{ij}^{\text{spa}}, ~D_{ii}^{\text{spe}} = \sum_{j=1}^{n} W_{ij}^{\text{spe}}.
\end{equation}

 \subsection{Multi-Order Graph}
 
Although constructing first-order nearest neighbor graphs can achieve a low-dimensional approximation of adjacent pixel representations \cite{liu2024towards}, such local connectivity constraints often fail to characterize the global information inherent in HSIs, where the similarity between many pixels may extend beyond direct neighbors. In this regard, high-order graphs can be considered to improve the accuracy of HU \cite{wu2022adaptive}. Let $\mathbf{W} \in \mathbb{R}^{N \times N}$ denote the first-order graph with $N$ nodes, and the $K$-order graph \( \mathbf{W} _K \) is defined as 
\begin{equation}\label{highorder}
\mathbf{W} _K =
\begin{cases}
\mathbf{W}, & K = 1, \\
\mathbf{W} _{K-1} \mathbf{W}, & K > 1.
\end{cases}
\end{equation}

After obtaining \( \mathbf{W}_K \), the corresponding Laplacian matrix can be calculated by

\begin{equation}
\mathbf{L}_K = \mathbf{D}_K - \mathbf{W}_K,
\end{equation}
where \( D_{Kii} = \sum_{j=1}^{N} W_{Kij} \). The fused higher-order Laplacian matrix \( \mathbf{L}_m \), degree matrix \( \mathbf{D}_m \), and weight matrix \(\mathbf{W}_m \) are 
\begin{equation}
	\begin{aligned}
\mathbf{W}_m &= \varrho _1 \mathbf{W}_1 + \varrho _2 \mathbf{W}_2 + \dots + \varrho _K \mathbf{W}_K,\\
\mathbf{D}_m &= \varrho _1 \mathbf{D}_1 + \varrho _2 \mathbf{D}_2 + \dots + \varrho _K \mathbf{D}_K,		\\
\mathbf{L}_m &= \varrho _1 \mathbf{L}_1 + \varrho _2\mathbf{L}_2 + \dots + \varrho _K \mathbf{L}_K,		
	\end{aligned}
\end{equation}
where  \( \varrho _1, \dots, \varrho _k \) are the weight  parameters.
From both spatial and spectral perspectives, we construct the following spatial and spectral multi-order graph

\begin{equation}
	\begin{aligned}
\mathbf{W}_m &= \varrho_1 \mathbf{W}_1^{\text{spa}} + \varrho _2 \mathbf{W}_2^{\text{spa}} + \dots + \varrho _K \mathbf{W}_K^{\text{spa}} \\
&~~~+ \eta_1 \mathbf{W}_1^{\text{spe}} + \eta_2 \mathbf{W}_2^{\text{spe}} + \dots + \eta_K \mathbf{W}_K^{\text{spe}},\\
\mathbf{D}_m& =\varrho _1 \mathbf{D}_1^{\text{spa}} + \varrho _2 \mathbf{D}_2^{\text{spa}} + \dots + \varrho _K \mathbf{D}_K^{\text{spa}} \\
&~~~+ \eta _1 \mathbf{D}_1^{\text{spe}} + \eta _2 \mathbf{D}_2^{\text{spe}} + \dots + \eta _K \mathbf{D}_K^{\text{spe}},\\
\mathbf{L}_m &=\varrho _1 \mathbf{L}_1^{\text{spa}} + \varrho _2 \mathbf{L}_2^{\text{spa}} + \dots + \varrho _K \mathbf{L}_K^{\text{spa}} \\
&~~~+ \eta_1 \mathbf{L}_1^{\text{spe}} + \eta_2 \mathbf{L}_2^{\text{spe}} + \dots + \eta_K \mathbf{L}_K^{\text{spe}},
	\end{aligned}
\end{equation}
where  \( \eta _1, \dots, \eta _K \) are also the weight  parameters. Furthermore, if two pixels exhibit similarity both spatially and spectrally, their abundances should also be similar. Therefore, we propose minimizing the following objective function to characterize this relationship
\begin{equation}\label{parameter}
	\begin{aligned}
&\frac{1}{2} \sum_{i,j=1}^{N} \| \mathbf{s}_i - \mathbf{s}_j \|_2^2 W_{mij} \\
&= \sum_{i=1}^{N} \mathbf{s}_i^\top \mathbf{s}_i D_{mij} - \sum_{i,j=1}^{N} \mathbf{s}_i^\top \mathbf{s}_j W_{mij} \\
&= \text{Tr}(\mathbf{S} \mathbf{D}_m \mathbf{S}^\top) - \text{Tr}(\mathbf{S} \mathbf{W}_m \mathbf{S}^\top) \\
&= \text{Tr}(\mathbf{S} \mathbf{L}_m \mathbf{S}^\top),
	\end{aligned}
\end{equation}
where $\mathbf{L}_m$ is the adaptive Laplacian matrix. Compared to traditional $k$-NN or heat kernel-based graph constructions, this adaptive Laplacian matrix can be dynamically adjusted according to different hyperspectral data, thereby capturing local and global information more comprehensively. In addition, it can learn different order graphs in the spectral and spatial domains, which helps to filter out irrelevant connections caused by noise or sensor defects, making the graph structure more stable.

\subsection{Our Model}

In this paper, we introduce multi-order graphs and robust LMM into the NMF framework, given by
\begin{equation}\label{main}
	\begin{aligned}
		\min_{\mathbf{A}, \mathbf{S}, \mathbf{E} } \quad &\frac{1}{2} \| \mathbf{X}-\mathbf{E}  - \mathbf{A} \mathbf{S} \|_F^2 + \gamma \| \mathbf{S} \|_{1/2} \\
		& \quad + \beta \| \mathbf{E} \|_{2,1} + \frac{\lambda}{2} \text{Tr}(\mathbf{S} \mathbf{L}_m \mathbf{S}^\top)  \\	
		\textrm{s.t.} \quad~~& \mathbf{A} \geq 0, ~\mathbf{S}  \geq 0,~ \mathbf{1}_{M}^{\top } \mathbf{S} = \mathbf{1}_{N}^{\top},
	\end{aligned}
\end{equation}
where $\gamma, \beta,  \lambda$ are the regularization parameters. Compared with HGNMF-FS in \eqref{HGNMF-FS}, the advantages of our proposed model  are  as follows.
\begin{itemize}
\item $\text{Tr}(\mathbf{S} \mathbf{L}_m \mathbf{S}^\top) $ exploits multi-order graphs to capture the global and local structures of HSIs.
\item  The weight parameters associated with $\mathbf{L}_m$ can be tuned adaptively, which will be discussed in Subsection \ref{ada-para}.
	\item $\| \mathbf{E} \|_{2,1} $ is enforced to filter out sparse noise and corruptions, thus improving robustness.
\end{itemize}

Note that our proposed model in \eqref{main} can also involve $\mathbf{\Lambda}$ to achieve adaptive feature selection, but this is not the focus of the current study.

\section{Optimization Algorithm}\label{Optimization Algorithm}

In this section, we provide the update scheme for solving 	\eqref{main} and the learning strategy for parameters in \eqref{parameter}.

\subsection{Update Scheme}

According to the alternating minimization technique, all the variables  $\mathbf{A}, \mathbf{S}, \mathbf{E}$ can be optimized one by one, which will be described in detail.

\subsubsection{Update \(\mathbf{A}\)} 
By applying the Lagrangian method, the subproblem can be characterized by
\begin{equation}
\begin{aligned}\label{lagrange1}
\Gamma  ( \mathbf{A}, \mathbf{Y}  ) &= \frac{1}{2} \| \mathbf{X} - \mathbf{E}  - \mathbf{A} \mathbf{S} \|_F^2 + \text{Tr}( \mathbf{Y} \mathbf{A}^\top),
\end{aligned}  
\end{equation}
where \( \mathbf{Y} \) is the Lagrangian multiplier.

The derivative of (\ref{lagrange1}) with respect to \( \mathbf{A} \) is
\begin{equation}
	\frac{\partial \Gamma ( \mathbf{A}, \mathbf{Y}  ) }{\partial \mathbf{A}} = -(\mathbf{X} - \mathbf{E} ) \mathbf{S}^{\top} + \mathbf{A} \mathbf{S}  \mathbf{S}^{\top} +\mathbf{Y}.
\end{equation}

From the Karush-Kuhn-Tucker (KKT) conditions, it holds
\begin{equation}
	\left( -(\mathbf{X} - \mathbf{E}) \mathbf{S}^\top + \mathbf{A} \mathbf{S} \mathbf{S}^\top \right) .* \mathbf{A} = 0,
\end{equation}
which shows that the solution  for \( \mathbf{A} \) can be expressed by 

\begin{equation}
\mathbf{A} \leftarrow \mathbf{A} .* ((\mathbf{X} - \mathbf{E}) \mathbf{S}^\top)\cdot /(\mathbf{A} \mathbf{S} \mathbf{S}^\top).
	\label{updateA}
\end{equation}

\subsubsection{Update  \(\mathbf{S}\)} 

The corresponding Lagrangian function can be constructed as
\begin{equation}
\begin{aligned}
\Psi (\mathbf{S}, \mathbf{Z}) &= \frac{1}{2} \| \mathbf{X} - \mathbf{E} - \mathbf{A} \mathbf{S} \|_F^2 + \gamma \| \mathbf{S} \|_{1/2}  \\
&~~~+ \frac{\lambda}{2} \text{Tr}(\mathbf{S}^\top \mathbf{L}_m \mathbf{S})+ \text{Tr}(\mathbf{Z} \mathbf{S}^\top),
\end{aligned}
\label{lagrange2}
\end{equation}
where \( \mathbf{Z}  \) is the Lagrangian multiplier.

Taking the derivative of (\ref{lagrange2}) with respect \(\mathbf{S}\), it has
\begin{equation}
	\begin{aligned}
	\frac{\partial 	\Psi (\mathbf{S}, \mathbf{Z}) }{\partial \mathbf{S}} &= -\mathbf{A}^{\top} (\mathbf{X} - \mathbf{E}) + \mathbf{A}^{\top} \mathbf{A}\mathbf{S} \\
&~~~+ \frac{\gamma}{2} \mathbf{S}^{-1/2} + \lambda  \mathbf{S} \mathbf{L}_m + \mathbf{Z}.
	\end{aligned}
\end{equation}

From the KKT conditions, it obtains
\begin{align}
    &\left( -\mathbf{A}^{\top} (\mathbf{X} - \mathbf{E}) 
    + \mathbf{A}^{\top} \mathbf{A}\mathbf{S} \right. \notag \\ 
    &\quad\quad\quad \left. + \frac{\gamma}{2} \mathbf{S}^{-1/2} 
    + \lambda \mathbf{S} \mathbf{L}_m \right) .* \mathbf{S} = 0.
\end{align}
This, together with the fact that \( \mathbf{L}_m = \mathbf{D}_m - \mathbf{W}_m \), derives the solution as follows
\begin{equation}
	\begin{aligned}\label{updateS}
\mathbf{S} \leftarrow \mathbf{S} &. *\left(\mathbf{A}^{\top} (\mathbf{X} - \mathbf{E}) + \lambda \mathbf{S} \mathbf{W}_m\right) \cdot /  \left(\mathbf{A}^{\top} \mathbf{A} \mathbf{S}\right. \\
& \left.+\frac{\gamma}{2}  \mathbf{S}^{-1/2}+\lambda \mathbf{S} \mathbf{D}_m\right) .
	\end{aligned}
\end{equation}

\subsubsection{Update  \(\mathbf{E}\)} 
After updating  \(\mathbf{A}\) and  \(\mathbf{S}\), the  subproblem associated with  \(\mathbf{E}\) is
\begin{equation}	
		\min_{\mathbf{E} } ~\frac{1}{2} \| \mathbf{X}-\mathbf{E}  - \mathbf{A} \mathbf{S} \|_F^2 +\beta \| \mathbf{E} \|_{2,1},
\end{equation}
which admits the following closed-form solution
\begin{equation}
	\mathbf{E} \leftarrow \textit{soft}_{\beta}(\mathbf{X} - \mathbf{A}  \mathbf{S} ).
	\label{updateE}
\end{equation}
For ease of expression, let $\mathbf{T}=\mathbf{X} - \mathbf{A}  \mathbf{S} $. The above soft thresholding  operator is defined as
\begin{equation}	
\textit{soft}_{\beta}(\mathbf{t}^i)=
\begin{cases}
\frac{\| \mathbf{t}^i \|_2 - \beta}{\| \mathbf{t}^i \|_2} \mathbf{t}^i, & \text{if} \ \| \mathbf{t}^i \|_2 \geq \beta, \\
0, & \text{otherwise},
\end{cases}
\end{equation}
where $ \mathbf{t}^i$ is the $i$-th row of $\mathbf{T}$, and $\beta$ is the threshold. See \cite{liu2012robust}  for more illustrations.

Finally, Algorithm \ref{MOGNMF} summarizes the overall framework for solving (\ref{main}), where the parameters of multi-order graphs in updating \( \mathbf{S} \) can be adaptively learned through Algorithm \ref{getgraph}. Here, \( \mathcal{L}_1^{(i)}\) denotes the objective value after the \(i\)-th update.

\begin{algorithm}[t]
    \caption{The procedure for solving (\ref{main})}
    \label{MOGNMF}
    \textbf{Input:} Data matrix \( \mathbf{X} \in \mathbb{R}^{L \times N} \), parameters \( \mu, \alpha, \beta, \gamma, \lambda \) \\
    \textbf{Initialize:} $(\mathbf{A}^0, \mathbf{S}^0, \mathbf{E}^0), \; \epsilon_1 = 10^{-4}, \; T_1 = 3000$ \\
    \textbf{Repeat}
    \begin{algorithmic}[1]
        \STATE Update  \( \mathbf{A} \) by (\ref{updateA})
        \STATE Update \( \mathbf{S} \) by (\ref{updateS})
        \STATE Update \( \mathbf{E} \) by (\ref{updateE})
        \STATE Compute objective value
        \[
        \mathcal{L}_1^{(i)} = \|\mathbf{X} - \mathbf{A}^{(i)}\mathbf{S}^{(i)} \|_F^2
        \]
       \STATE Set \( i \leftarrow i + 1 \)
    \end{algorithmic}
    \textbf{Until} \( |\mathcal{L}_1^{(i)} - \mathcal{L}_1^{(i-1)}| < \epsilon_1 \) \textbf{or} \( i \geq T_1 \) \\
    \textbf{Output:} $(\mathbf{A}, \mathbf{S}, \mathbf{E})$
\end{algorithm}

\subsection{Parameter Adaptive Learning} \label{ada-para}

To learn \(\mathbf{W}_{m} \), the following model is proposed
\begin{equation}
	\begin{aligned}
		\label{eq:zong}
		\min_{\mathbf{H}, \mathbf{W}_{m}} \quad & \sum_{v=1}^{V} \sum_{k=1}^{K} H_{vk} \|\mathbf{W}_{m} - \mathbf{W}_k^v \|_F^2 	\\
		& 	+ \mu \|\mathbf{W}_{m} \|_F^2 +  \alpha  \|\mathbf{H} \|_F^2 \\
		\textrm{s.t.} ~\quad & \mathbf{W}_{m} \geq 0, \, \mathbf{1}_V^\top \mathbf{H} \mathbf{1}_K = 1, \, \mathbf{H} \geq 0,
	\end{aligned}
\end{equation}
of which \(\mathbf{W}_k^v \) represents the \( k \)-order graph of the spatial or spectral graph.
\( \mathbf{H} \in \mathbb{R}^{V \times K} \) is the coefficient matrix, where \( H_{vk} \) denotes the corresponding weight of \(\mathbf{W}_k^v\) in the consensus graph \( \mathbf{W}_m \). \(V\) represents the spatial or spectral graph, with \(v=1\) representing the spatial graph and \(v=2\) representing the spectral graph.  In addition, $\mu$ and $\alpha$ are the regularization parameters to prevent overfitting.

\subsubsection{Update \( \mathbf{W}_{m} \)}

The solution can be simplified as 
\begin{equation}
\begin{aligned}
	\min_{\mathbf{W}_{m}} \quad &\sum_{v=1}^{V} \sum_{k=1}^{K} H_{vk} 
 	\left( \text{Tr}(\mathbf{W}_{m}^\top \mathbf{W}_{m}) - 2 \text{Tr}(\mathbf{W}_{m}^\top \mathbf{W}_k^v) \right) \\
	&+ \mu \|\mathbf{W}_{m} \|_F^2\\
    \textrm{s.t.} \quad & \mathbf{W}_{m} \geq 0.
\end{aligned}
\label{eq:zong1} 
\end{equation}

Denote the objective as $g(\mathbf{W}_{m})$. The gradient  is
\begin{equation}
	\nabla {g(\mathbf{W}_m)}  = 2 \left( 1 + \mu \right) \mathbf{W}_m - 2 \sum_{v=1}^{V} \sum_{k=1}^{K} H_{vk} \mathbf{W}_k^v.
	\label{zong2}
\end{equation}

Setting the gradient to zero and considering the constraint $\mathbf{W}_{m} \geq 0$, the results can be obtained as
\begin{equation}
  \mathbf{W}_m \leftarrow \max \left( 0,  \sum_{v=1}^{V} \sum_{k=1}^{K} H_{vk} \mathbf{W}_k^v / \left( 1 + \mu \right) \right).
    \label{UPDATEW}
\end{equation}

\subsubsection{Update \( \mathbf{H} \) } Let  \( P_{vk} = \| \mathbf{W}_{m} - \mathbf{W}_k^v \|_F^2 \). Then the subproblem can be written as
\begin{equation}
\begin{aligned}
    \min_{\mathbf{H}} \quad & \sum_{v=1}^{V} \sum_{k=1}^{K} H_{vk} P_{vk} + \alpha \| \mathbf{H} \|_F^2 \\
    \textrm{s.t.} \quad & \mathbf{1}_V^\top \mathbf{H} \mathbf{1}_K = 1, \, \mathbf{H} \geq 0.
\end{aligned}
\label{eq:zong3}
\end{equation}
Further, let \( \hat{\mathbf{H}} = \text{vec}(\mathbf{H}) \) and \( \hat{\mathbf{P}} = \text{vec}(\mathbf{P}) \). Therefore, the above problem can be reformulated as
\begin{equation}
	\min_{\hat{\mathbf{H}}} \quad \alpha \hat{\mathbf{H}}^\top \hat{\mathbf{H}} + \hat{\mathbf{P}}^\top \hat{\mathbf{H}},
	\label{eq:zong4}
\end{equation}
which can be solved by existing quadratic programming (QP) solvers. 

After obtaining \( \mathbf{W}_m \) through Algorithm \ref{getgraph}, \( \mathbf{L}_m \) and \( \mathbf{D}_m \) can be computed using \( D_{mii} = \sum_{j=1}^{N} W_{mij} \) and \( \mathbf{L}_m = \mathbf{D}_m - \mathbf{W}_m \), which are then used in Algorithm \ref{MOGNMF}.

 \begin{algorithm}[t]
	\caption{The procedure for solving (\ref{eq:zong})}  
	\label{getgraph}
	\textbf{Input:} Order \( K \), parameters \( \mu \), \( \alpha \) \\
	\textbf{Initialize:} \( H_{vk} = \frac{1}{2K}, \epsilon_2  = 10^{-6}, \; T_2 = 50 \), multi-order graphs via (\ref{highorder})\\
	 \textbf{Repeat}
	\begin{algorithmic}[1]
		\STATE  Update \( \mathbf{W}_{m} \) by (\ref{UPDATEW})
		\STATE  Update \( \mathbf{H} \) by (\ref{eq:zong4})
        \STATE Compute objective value:
			\[
			\mathcal{L}_2^{(j)} = \sum_{v=1}^{V} \sum_{k=1}^{K} H_{vk}^{(j)}  \|\mathbf{W}_{m}^{(j)}  - \mathbf{W}_k^v \|_F^2 + \mu \|\mathbf{W}_{m}^{(j)}  \|_F^2 +  \alpha  \|\mathbf{H}^{(j)}  \|_F^2
			\]
			\STATE Set \( j \leftarrow j + 1 \)
	\end{algorithmic}
	\textbf{Until} \( |\mathcal{L}_2^{(j)} - \mathcal{L}_2^{(j-1)}| < \epsilon_2 \) \textbf{or} \( j \geq T_2 \) \\
			\textbf{Output:}   \( (\mathbf{W}_{m},  \mathbf{H}) \) 
\end{algorithm}

\begin{Remark}
We would like to point out that: (1) For the \( \mathbf{S} \)-subproblem, $\ell_{1/2}$-norm is introduced to promote sparsity in the abundance maps, but it is nonsmooth and nonconvex. (2) For the \( \mathbf{E} \)-subproblem, $\ell_{2,1}$-norm is incorporated to enhance robustness against noise and outliers, but it is nonsmooth yet convex. (3)  In the experimental part, the proposed algorithm is validated to exhibit a monotonically decreasing objective value during the optimization process.
\end{Remark}

\subsection{Computational Complexity Analysis} \label{Computation-ana}
\subsubsection{Time Complexity}

In Algorithm~\ref{MOGNMF}, the computational complexity of standard NMF is $\mathcal{O}(L N M )$. The update of $\mathbf{E}$ involves applying soft-thresholding to each row, with a complexity of $\mathcal{O}(L N)$. In addition, updating $\mathbf{S}$ requires computing $\mathbf{S}^{-1/2}$, which incurs a computational cost of $\mathcal{O}(M N)$. The computation of $\mathbf{S} \mathbf{W}_m$ and $\mathbf{S} \mathbf{D}_m$ both incur a complexity of $\mathcal{O}(M N^2)$. 
Therefore, the overall computational complexity of Algorithm~\ref{MOGNMF} is given by 
\begin{equation}
\begin{aligned}
\mathcal{O}(&L N M +L N + M N+ M N^2).
\end{aligned}
\end{equation}

In Algorithm~\ref{getgraph}, the computational complexity of updating $\mathbf{W}_{m}$ is $\mathcal{O}(2K N^2)$. Updating $\mathbf{H}$ involves solving a QP problem, whose complexity is $\mathcal{O}((V K)^3)$. Since $V$ and $K$, representing the number of views and the graph order respectively, are relatively small, the cost of updating $\mathbf{H}$ can be considered negligible. 
It is worth noting that the algorithm requires constructing first-order graphs from both spectral and spatial views. We use nearest neighbour $C$, and the corresponding C-nearest neighbor graphs have a computational cost of $\mathcal{O}(N^2 L)$ and $\mathcal{O}(N^2)$, respectively. For each higher-order graph, the second-order weight matrix incurs a cost of $\mathcal{O}(C^2 N)$. Therefore, by recursive reasoning, the total complexity for constructing $K$-order graphs can be derived as
\begin{equation}
\sum_{i=2}^{K} \mathcal{O}(C^2 N) = (K - 1) \cdot \mathcal{O}(C^2 N) = \mathcal{O}(K C^2 N).
\end{equation}
Therefore, the overall computational complexity of Algorithm~\ref{getgraph} is given by 
\begin{equation}
\mathcal{O}(2K N^2+N^2 L+N^2+K C^2 N).
\end{equation}

If Algorithm~\ref{MOGNMF} terminates after $\psi$ iterations, and Algorithm~\ref{getgraph} terminates after $\upsilon $ iterations, then the total computational cost can be expressed as  
\begin{equation}
\begin{aligned}
\mathcal{O}(&\psi (L N M +L N + M N) + \upsilon  (2K N^2)\\
& +N^2 L + N^2 + K C^2 N+M N^2).
\end{aligned}
\end{equation}

\subsubsection{Space Complexity}
The space complexity of MOGNMF is determined by the memory usage of the data matrix $\mathbf{X} \in \mathbb{R}^{L \times N}$, the endmember matrix $\mathbf{A} \in \mathbb{R}^{L \times M}$, the abundance matrix $\mathbf{S} \in \mathbb{R}^{M \times N}$, the noise matrix $\mathbf{E} \in \mathbb{R}^{L \times N}$, the fused graph matrix $\mathbf{W}_m \in \mathbb{R}^{N \times N}$, and the multi-order weight matrix $\mathbf{H} \in \mathbb{R}^{2K \times 2K}$. 
Therefore, the overall space complexity is given by
$\mathcal{O}(LN + LM + MN + N^2 + 4K^2).$
Since $K \ll N$, the term $4K^2$ can be considered negligible. Thus, the overall space complexity simplifies to
\begin{equation}
\begin{aligned}
\mathcal{O}(LN + LM + MN + N^2).
\end{aligned}
\end{equation}

\section{Numerical Experiments}\label{Numerical Experiments}

This section presents numerical experiments between our method and some state-of-the-art HU methods, including VCA-FCLS\footnote{\url{https://github.com/isaacgerg/matlabHyperspectralToolbox}} \cite{nascimento2005vertex} \cite{heinz2001fully}, MVC-NMF\footnote{\url{https://github.com/aicip/MVCNMF}} \cite{miao2007endmember}, SNMF\footnote{\url{https://github.com/yuehniu/Remote.Sensing}} \cite{qian2011hyperspectral} , RNMF\footnote{\url{https://www.irit.fr/~Cedric.Fevotte/extras/tip2015/code.zip}}\cite{fevotte2015nonlinear} , SGSNMF\footnote{\url{https://github.com/YW81/TGRS17-SGSNMF}}  \cite{wang2017spatial}, IRNMF \cite{zhang2024matrix}, HGNMF-FS \cite{qu2023hyperspectral}, MV-DUAL\footnote{\url{https://github.com/mabdolali/MaxVol_Dual/}}  \cite{abdolali2024dual}, and A2SAN\footnote{\url{https://github.com/xuanwentao/A2SN-and-A2SAN}} \cite{tao2024abundance}.

Subsection \ref{Experimental Setup} outlines the experimental setup. Subsection \ref{Simulated Datasets}  introduces the simulated datasets and their results, while Subsection \ref{Results on Real-world Datasets}  discusses the real datasets and their results. Subsection \ref{Ablation Studies} provides the ablation experiments. Subsection \ref{Discussion} offers further discussions.

\begin{figure}[t]
		\centering
		\subfigcapskip=-3pt
		\subfigure[]{
			\label{dc21}
			\centering
			\includegraphics[width=0.2\textwidth]{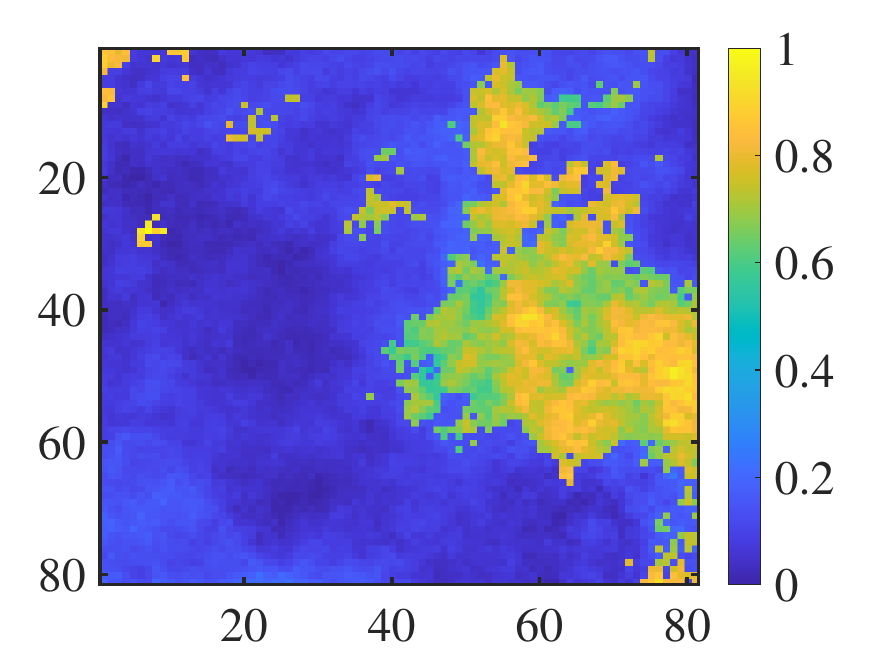}
		}
		\subfigcapskip=-3pt
		\subfigure[]{
			\label{dc22}
			\centering
			\includegraphics[width=0.2\textwidth]{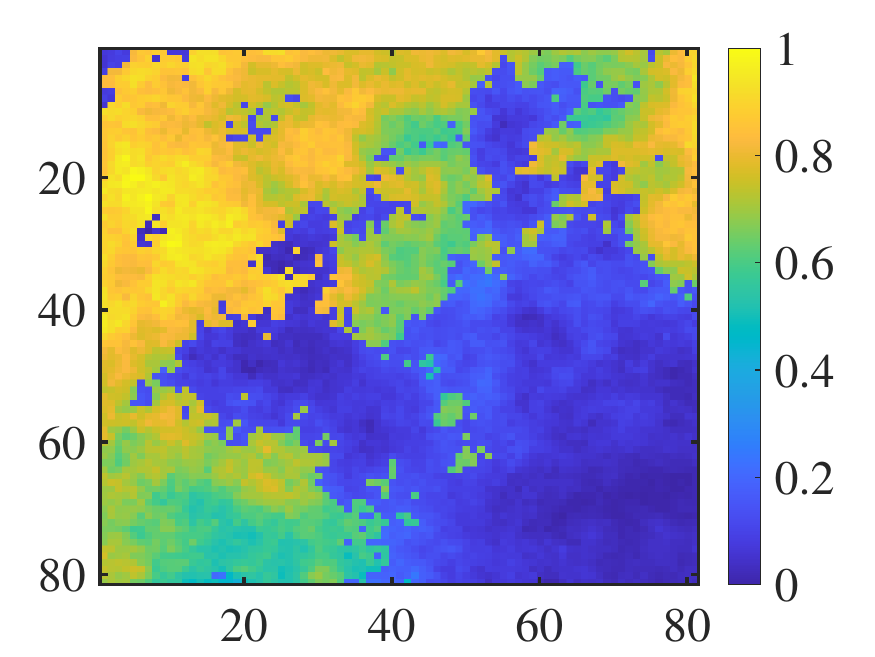}
		}
		\\ 
		\subfigcapskip=-3pt
		\subfigure[]{
			\label{dc23}
			\centering
			\includegraphics[width=0.2\textwidth]{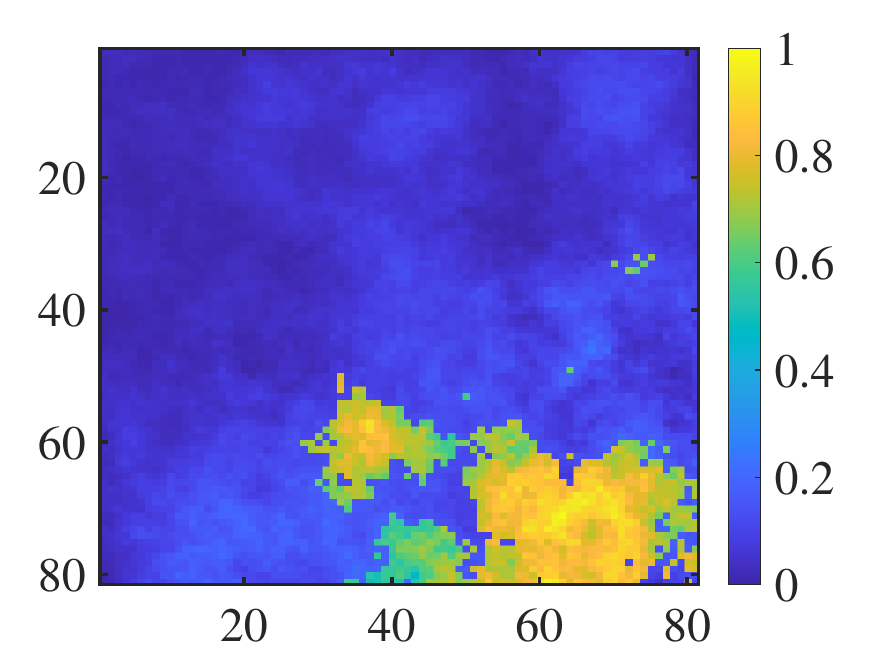}
		}
		\subfigcapskip=-3pt
		\subfigure[]{
			\label{dc24}
			\centering
			\includegraphics[width=0.2\textwidth]{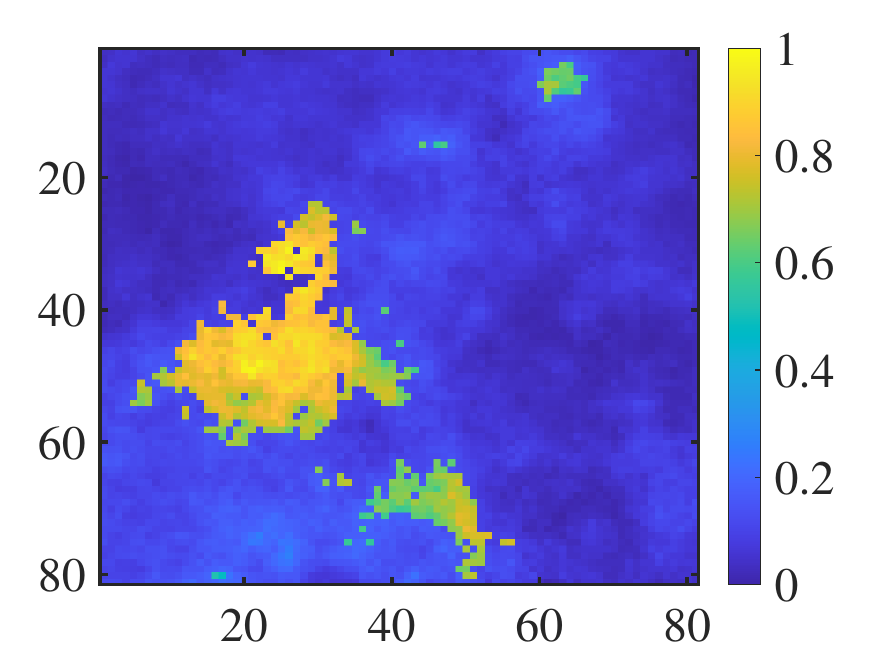}
		}
		\vspace{-0.2cm}
		\caption{Illustration of the Simu-1 dataset, where (a)-(d) are the abundance maps in the case of $M = 4$.}
		\label{DC1}
	\end{figure}
	
\begin{figure}[t] 
    \centering
    \setlength{\abovecaptionskip}{-3pt} 
    \subfigure[]{
        \label{DC2a}
        \centering
        \includegraphics[width=0.2\textwidth]{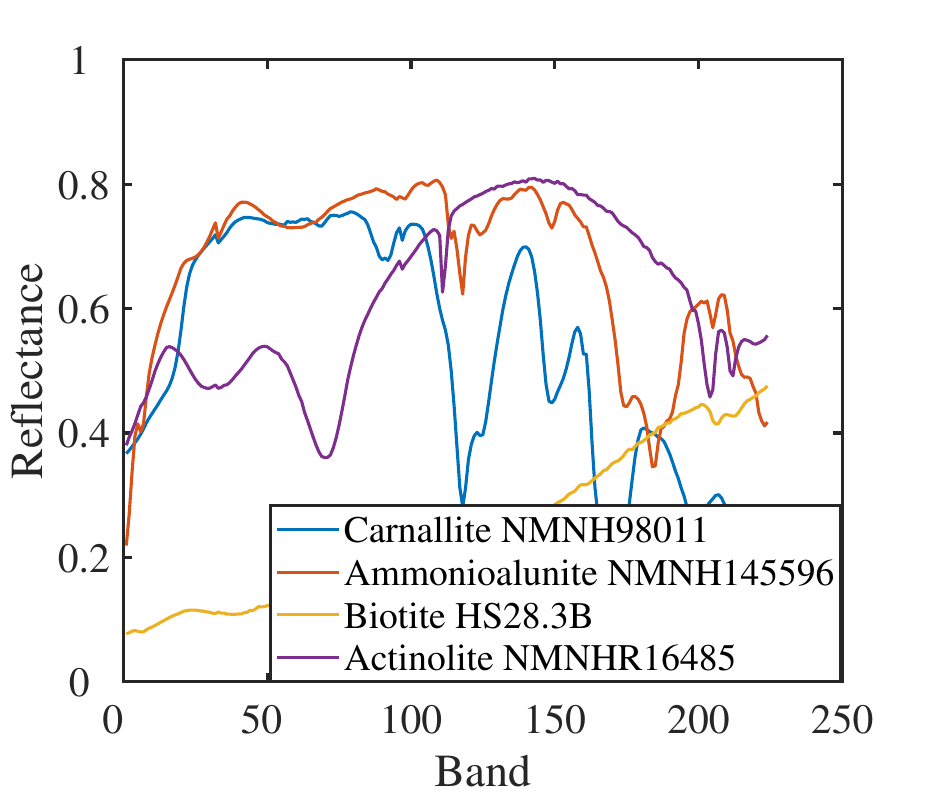} 
    }
    \subfigure[]{
        \label{DC2b}
        \centering
        \includegraphics[width=0.25\textwidth]{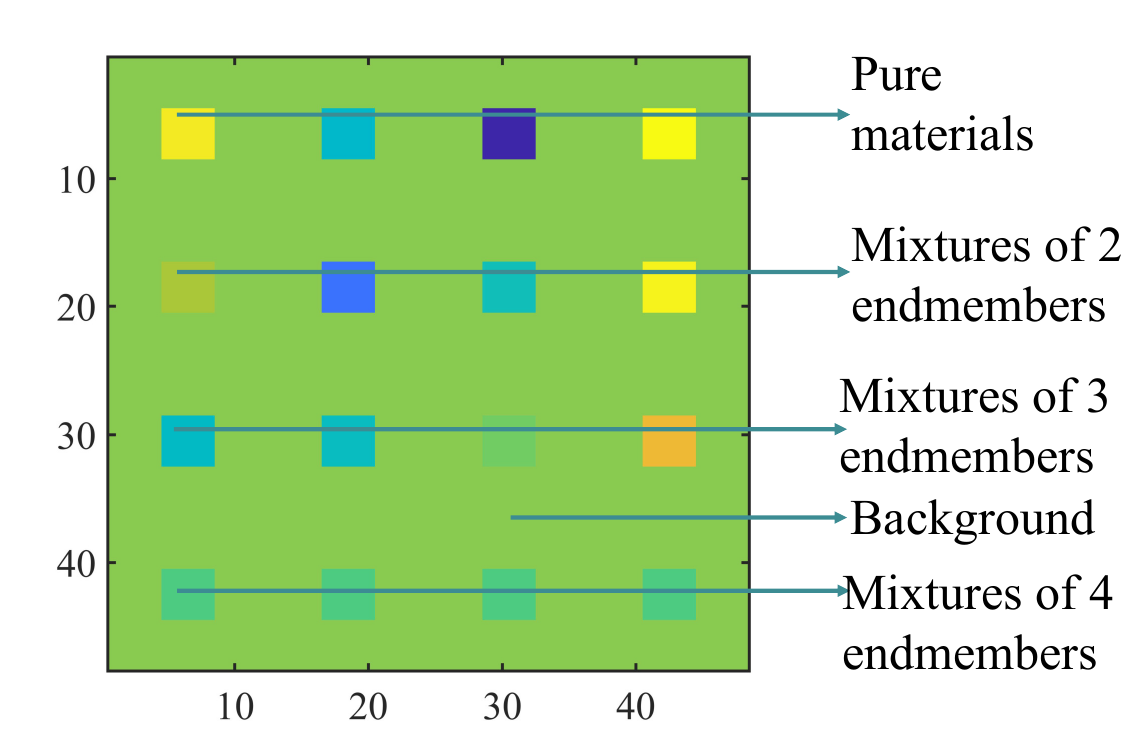} 
    }
    \vspace{-0.2cm}
    \caption{Illustration of the Simu-2 dataset, where (a) is the spectral signatures of selected endmembers and (b) is the simulated hyperspectral image.}
    \label{DC2}
\end{figure}

\begin{figure*}[t]  
	\centering
     \vspace{-2cm}
    \includegraphics[width=1\textwidth, keepaspectratio]{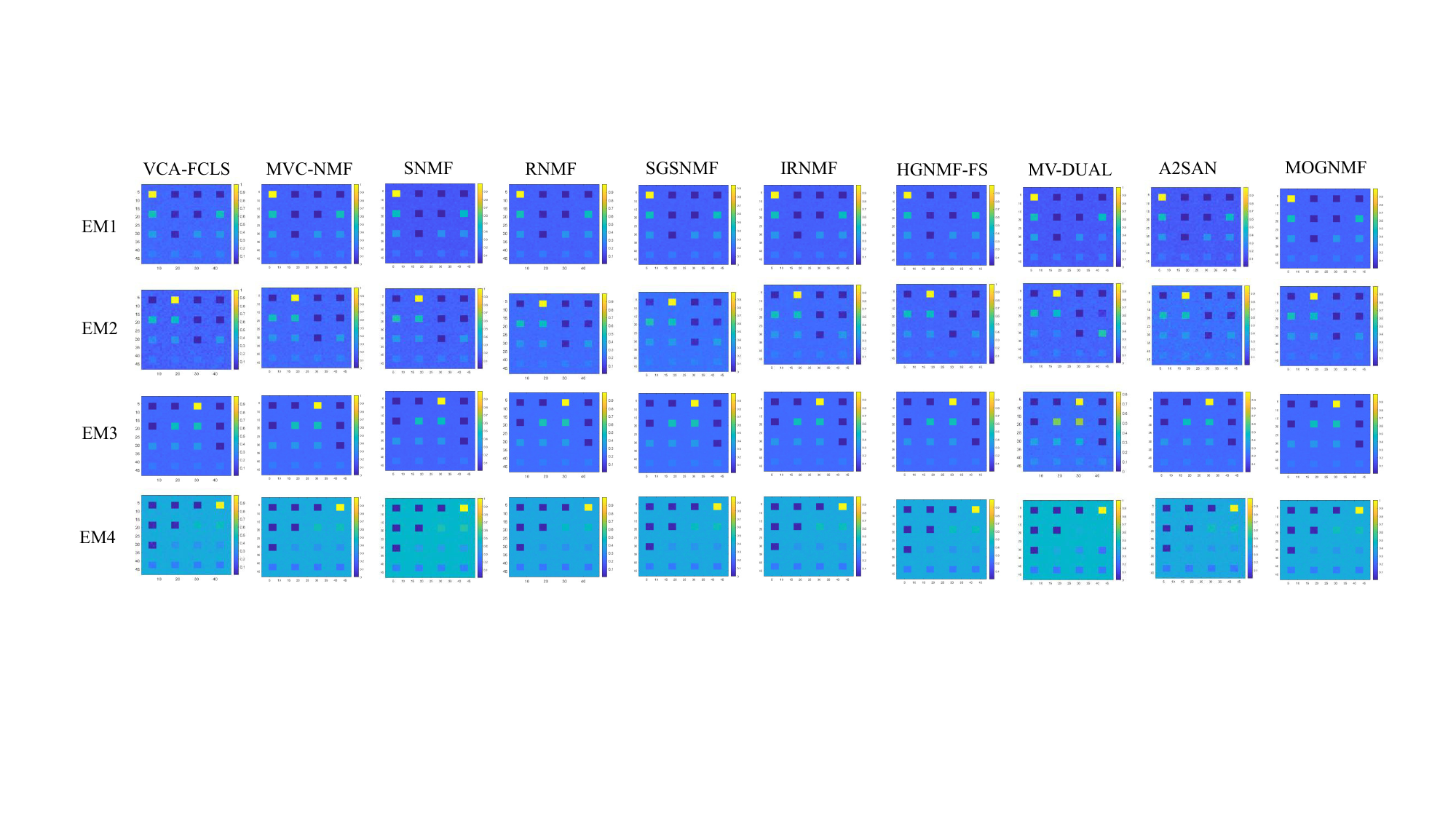}  
	\vspace{-3.6cm}
	\caption{Abundance maps obtained by the different unmixing methods on the Simu-2 dataset.}
	\label{DC2EM}
\end{figure*}

\begin{figure*}[t]  
	\centering
       \vspace{-1.4cm}
      \hspace*{-1.1cm}
	\includegraphics[width=1\textwidth, keepaspectratio]{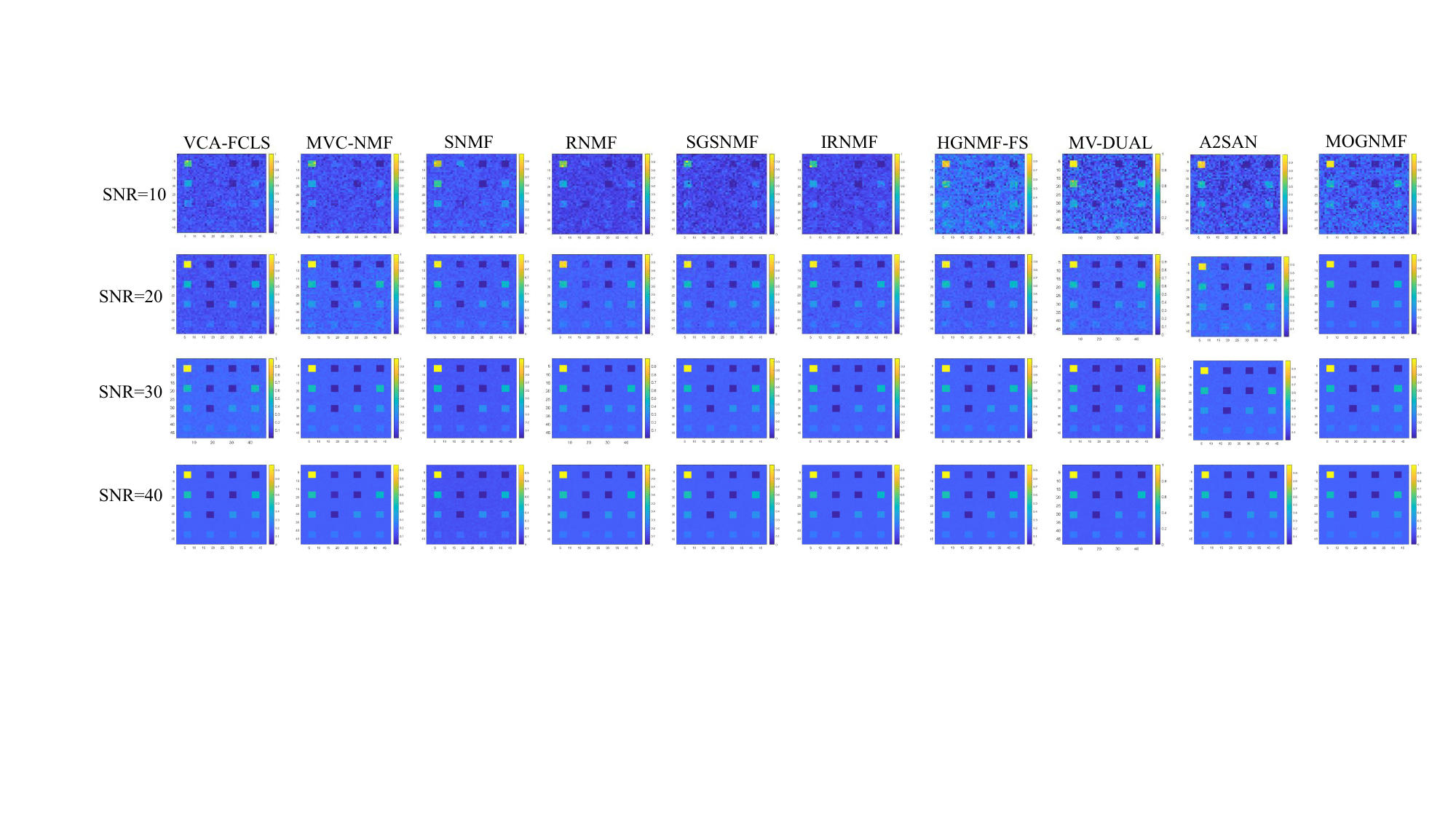}  
	\vspace{-3.6cm}
	\caption{Abundance maps of the endmember 1  obtained under different SNRs on the Simu-2 dataset.}
	\label{DC2SNR}
\end{figure*}

\subsection{Experimental Setup}\label{Experimental Setup}
\subsubsection{Parameter Settings}
For our proposed MOGNMF,  $\gamma$,  $\beta$, and $\lambda$  should be taken into account. 
According to \cite{qian2011hyperspectral}, $\gamma$ is estimated by
\begin{equation}\label{gamma}
	\gamma = \frac{1}{\sqrt{L}} \sum_{l=1}^{L} \frac{\sqrt{N - 1} \left\| \mathbf{x}^l \right\|_1}{\left\| \mathbf{x}^l \right\|_2 \sqrt{N - 1}},
\end{equation}
where \( L \) represents the number of spectral bands, \( N \) represents the total number of pixels, and  
\( \mathbf{x}^l \) represents the vector corresponding to the \( l \)-th band in the hyperspectral data \( \mathbf{X} \in \mathbb{R}^{L \times N} \). The regularization parameters are selected from the candidate set $\{ 10^{-3}, 10^{-2}, 10^{-1}, 10^{0}, 10, 10^{2}, 10^{3} \}$. For other compared methods, either the default values or the best parameters recommended by the original authors are adopted.

\subsubsection{Initializations}

Due to the nonconvexity of problem \eqref{main}, the initialization of \( \mathbf{A} \) and \( \mathbf{S} \) significantly impacts the results. Inspired by the existing work, we use VCA-FCLS \cite{nascimento2005vertex} \cite{heinz2001fully} for initialization. Besides, it is important to satisfy the constraints that both matrices \( \mathbf{A} \) and \( \mathbf{S} \)  are nonnegative (also ANC) and the columns of \( \mathbf{S} \) should sum to one (also ASC). In fact, during the iterative process, as long as the initial matrices \( \mathbf{A} \) and \( \mathbf{S} \) are nonnegative, we can naturally maintain the nonnegativity. The ASC constraint can be implemented using the efficient and widely used method proposed in \cite{heinz2001fully}, where the data matrix \( \mathbf{X}-\mathbf{E} \) and the endmember matrix \( \mathbf{A} \) are augmented as
\begin{equation}
	\overline{\mathbf{X}-\mathbf{E}} = 
	\begin{pmatrix}
		\mathbf{X}-\mathbf{E} \\
		\delta \mathbf{1}_N^\top
	\end{pmatrix}, \quad
	\overline{\mathbf{A}} =
	\begin{pmatrix}
		\mathbf{A} \\
		\delta \mathbf{1}_M^\top
	\end{pmatrix},
	\label{augment}
\end{equation}
where \(\delta\) is used to control the effect of ASC. Larger $\delta$ values can improve the accuracy, but will also significantly reduce the convergence speed. To balance them, we set $\delta = 15$.

\subsubsection{Stopping Criteria}

For Algorithm \ref{MOGNMF}, two stopping criteria are considered. The first is based on a predefined error tolerance. Once the relative error is less than \(\epsilon_1 = 10^{-4}\), the iteration stops. The second  is the maximum number of iterations, which is set to 3000. The optimization process halts when either of these criteria is satisfied. For Algorithm \ref{getgraph}, the maximum number of iterations is 50. The iteration stops when either the maximum number of iterations is reached or the relative error is smaller than \(\epsilon_2=10^{-6}\).

\subsubsection{Evaluation Metrics}

To evaluate the HU performance of different methods, two popular metrics are employed. For the $k$-th estimated endmember vector $\mathbf{a}_k$ and its truth endmember $\hat{\mathbf{a}}_k$, the spectral angle distance (SAD) is defined as
\begin{equation}
\begin{aligned}
\text{SAD} = \cos^{-1} \left( \frac{\mathbf{a}_k^\top  \widehat{\mathbf{a}}_k}{\|\mathbf{a}_k\|_2 \|\widehat{\mathbf{a}}_k\|_2} \right).
\end{aligned}
\end{equation}
For the $k$-th estimated abundance vector $\mathbf{s}_k$ and  the ground truth abundance vector \( \hat{\mathbf{s}}_k \), the root mean square error (RMSE) is given by
\begin{equation}
	\begin{aligned}
 \text{RMSE} = \sqrt{\frac{1}{N} \sum_{k=1}^{N} \left\|  \mathbf{s}_k - \hat{\mathbf{s}}_k \right\|_2^2 }.
	\end{aligned}
\end{equation}
The smaller the SAD and RMSE values, the better the unmixing performance.

\subsection{Results on Simulated Datasets}\label{Simulated Datasets}

\subsubsection{Dataset Description} 

The Simu-1 dataset is constructed using the hyperspectral imagery synthesis toolbox\footnote{\url{http://www.ehu.es/ccwintco/index.php/Hyperspectral_Imagery_Synthesis_tools_for_MATLAB}}. This toolbox enables control over multiple key parameters, including spatial resolution, endmember spatial distribution patterns, and material composition characteristics. By randomly selecting a varying number of mineral spectra from the U.S. Geological Survey (USGS) digital spectral library as endmembers, where the number of endmembers is denoted as $M$, their spatial abundance distributions are modeled using Gaussian random fields with radial basis function covariance structures. Fig. \ref{DC1} presents the abundance maps when $M=4$.

The Simu-2 dataset is selected from the USGS digital repository\footnote{\url{http://speclab.cr.usgs.gov/spectral.lib06}}, containing reflectance data across 224 spectral channels (0.4-2.5 $\mu$m). This dataset is generated under the LMM framework with four materials, see Fig. \ref{DC2a}. The spatial layout includes discrete geometric regions of pure materials and complex mixtures, as shown in Fig. \ref{DC2b}.

\subsubsection{Numerical Results}

To emulate real acquisition conditions, the simulated hyperspectral cubes are corrupted with additive noise calibrated to specific signal-to-noise ratios (SNRs). The SNR quantification follows 
\begin{equation}
	\text{SNR} = 10 \log_{10} \frac{\mathbb{E}[\mathbf{x}^\top \mathbf{x}]}{\mathbb{E}[\mathbf{n}^\top \mathbf{n}]},
\end{equation}
where \( \mathbf{x} \) represents the noise-free signal vector, \( \mathbf{n} \) represents the noise vector, and \( \mathbb{E}[\cdot] \) denotes the mathematical expectation over all image pixels.

Tables \ref{DC1TABLE} and \ref{DC1TABLE2} provide the average SAD and RMSE results on the Simu-1 dataset contaminated by Gaussian white noise under SNR = 10, 20, 30, and  \( 40 \, \text{dB} \), respectively, where the top two values are marked as \textcolor[rgb]{1.00,0.00,0.00}{\textbf{red}} and \textcolor[rgb]{0.00,0.00,1.00}{\textbf{blue}}. It can be observed that when the SNR increases, the unmixing accuracy of all methods improves. Furthermore, compared with HGNMF-FS and RNMF, the proposed MOGNMF demonstrates superior performance under different SNR conditions. This indicates that the introduction of dual sparsity enhances robustness and also highlights the significant advantages of multi-order graph learning.

\begin{table*}[t]
	\renewcommand\arraystretch{1.2} 
	\caption{SAD values on the Simu-1 dataset under different SNR conditions, where the top two values are marked as \textcolor[rgb]{1.00,0.00,0.00}{\textbf{red}} and \textcolor[rgb]{0.00,0.00,1.00}{\textbf{blue}}. }\label{DC1TABLE}
		\vspace{-0.15cm}
	\centering
	\setlength{\tabcolsep}{1mm}  
	\begin{tabular}{|c|c|c|c|c|c|c|c|c|c|c|}
		\hline
	~ {\textbf{Cases}} ~~&~ \textbf{VCA-FCLS} ~ &~  {\textbf{MVC-NMF}}~  &~ {\textbf{SNMF}} ~ & ~ {\textbf{RNMF}} ~  &~ {\textbf{SGSNMF}} ~& ~  {\textbf{IRNMF}} ~ &~ {\textbf{HGNMF-FS}}~ & ~ {\textbf{MV-DUAL}} ~& ~ {\textbf{A2SAN}} ~&~{\textbf{MOGNMF}}~ \\ 
		\hline \hline
		SNR=10 & 0.0968&\textcolor[rgb]{0.00,0.00,1.00}{\textbf{0.0356}}&0.0442	&0.0627	&0.0358	&0.0361	&	0.0438		&0.0982&0.0514 & \textcolor[rgb]{1.00,0.00,0.00}{\textbf{0.0320}}  \\ \hline
		SNR=20  & 0.0821&	0.0350&	0.0524&	0.0610&	0.0436&	0.0340&		\textcolor[rgb]{0.00,0.00,1.00}{\textbf{0.0331}}&	0.0443&0.0421 & \textcolor[rgb]{1.00,0.00,0.00}{\textbf{0.0313}}  \\ \hline
	SNR=30& 0.0321&\textcolor[rgb]{0.00,0.00,1.00}{\textbf{0.0109}}&	0.0467&	0.0508&	0.0468&	0.0119&	0.0319&	0.0320&0.0259&\textcolor[rgb]{1.00,0.00,0.00}{\textbf{0.0099}}  \\ \hline
	SNR=40 & 0.0122&0.0090&	0.0460&	0.0149&	0.0393&	\textcolor[rgb]{0.00,0.00,1.00}{\textbf{0.0076}}&		0.0267&		0.0335&0.0108 &\textcolor[rgb]{1.00,0.00,0.00}{\textbf{0.0063}}  \\ \hline\hline
		Average & 0.0558& 0.0226& 0.0473& 0.0473&	 0.0432& \textcolor[rgb]{0.00,0.00,1.00}{\textbf{0.0224}}&	 0.0338& 0.0520& 0.0326& \textcolor[rgb]{1.00,0.00,0.00}{\textbf{0.0202}}\\ \hline
	\end{tabular}
\end{table*}

\begin{table*}[t]
	\renewcommand\arraystretch{1.2} 
	\caption{RMSE values on the Simu-1 dataset under different SNR conditions, where the top two values are marked as \textcolor[rgb]{1.00,0.00,0.00}{\textbf{red}} and \textcolor[rgb]{0.00,0.00,1.00}{\textbf{blue}}. }\label{DC1TABLE2}
		\vspace{-0.15cm}
	\centering
	\setlength{\tabcolsep}{1mm} 
	\begin{tabular}{|c|c|c|c|c|c|c|c|c|c|c|}
		\hline
	~ {\textbf{Cases}} ~~&~ \textbf{VCA-FCLS} ~ &~  {\textbf{MVC-NMF}}~  &~ {\textbf{SNMF}} ~ & ~ {\textbf{RNMF}} ~  &~ {\textbf{SGSNMF}} ~& ~  {\textbf{IRNMF}} ~ &~ {\textbf{HGNMF-FS}}~ & ~ {\textbf{MV-DUAL}} ~& ~ {\textbf{A2SAN}} ~&~{\textbf{MOGNMF}}~ \\ 
		\hline \hline
		SNR=10& 0.0943&	0.0966& 0.0988&0.0932&	0.0944&	0.0935&		0.1180&	0.1517&\textcolor[rgb]{0.00,0.00,1.00}{\textbf{0.0928}}& \textcolor[rgb]{1.00,0.00,0.00}{\textbf{0.0899}}  \\ \hline
		SNR=20 &\textcolor[rgb]{0.00,0.00,1.00}{\textbf{0.0924}}&	0.0937&	0.1008&	0.1151&	0.0941&	0.0955	&	0.0925&	0.1276&0.0981 & \textcolor[rgb]{1.00,0.00,0.00}{\textbf{0.0917}}  \\ \hline
		SNR=30 & 0.0552&\textcolor[rgb]{0.00,0.00,1.00}{\textbf{0.0310}}&	0.0894	& 0.0600& 0.0599 &0.0533&		0.0515&0.0394	&0.0475& \textcolor[rgb]{1.00,0.00,0.00}{\textbf{0.0276}}  \\ \hline
		SNR=40&0.0391	&\textcolor[rgb]{0.00,0.00,1.00}{\textbf{0.0319}}&0.0916	&0.0392&	0.0881&0.0376&		0.0619&	0.0462&0.0427 & \textcolor[rgb]{1.00,0.00,0.00}{\textbf{0.0219}}   \\ \hline\hline
		Average & 0.0703& \textcolor[rgb]{0.00,0.00,1.00}{\textbf{0.0633}}& 0.0952	& 0.0769&	 0.0841& 0.0700&	 0.0810& 0.0933&0.0683&  \textcolor[rgb]{1.00,0.00,0.00}{\textbf{0.0578}}\\ \hline
	\end{tabular}
\end{table*}

\begin{table*}[t]
	\renewcommand\arraystretch{1.2} 
	\caption{SAD comparisons on the Samon dataset, where the top two values are marked as \textcolor[rgb]{1.00,0.00,0.00}{\textbf{red}} and \textcolor[rgb]{0.00,0.00,1.00}{\textbf{blue}}.}\label{samonsad}
		\vspace{-0.15cm}
	\centering
	\setlength{\tabcolsep}{1mm} 
	\begin{tabular}{|c|c|c|c|c|c|c|c|c|c|c|}
		\hline
		~ {\textbf{Cases}} ~~&~ \textbf{VCA-FCLS} ~ &~  {\textbf{MVC-NMF}}~  &~ {\textbf{SNMF}} ~ & ~ {\textbf{RNMF}} ~  &~ {\textbf{SGSNMF}} ~& ~  {\textbf{IRNMF}} ~ &~ {\textbf{HGNMF-FS}}~ & ~ {\textbf{MV-DUAL}} ~& ~ {\textbf{A2SAN}} ~&~{\textbf{MOGNMF}}~ \\ 
		\hline \hline
		Soil& 0.0447&	0.0649&	\textcolor[rgb]{0.00,0.00,1.00}{\textbf{0.0129}}&	0.0286&	0.0209&	0.0166&	0.0237	&0.0265	&0.0686 & \textcolor[rgb]{1.00,0.00,0.00}{\textbf{0.0094}} \\ \hline
		Tree& 0.0434&	0.0532& \textcolor[rgb]{1.00,0.00,0.00}{\textbf{0.0351}}&	0.0716&	0.0443&	0.0374&	0.0419& 0.0386&	0.0422&
		\textcolor[rgb]{0.00,0.00,1.00}{\textbf{0.0368}}\\ \hline
		Water&0.1045&	0.1208&	0.1802&	0.1389&	0.0979&	0.1045&	0.0977& \textcolor[rgb]{1.00,0.00,0.00}{\textbf{0.0715}}&\textcolor[rgb]{0.00,0.00,1.00}{\textbf{0.0740}}&	0.0858 \\ \hline \hline
		Average & 0.0642&0.0796&	0.0761	&0.124&	0.0544&	0.0528&	0.0544&	\textcolor[rgb]{0.00,0.00,1.00}{\textbf{0.0455}}&0.0616& \textcolor[rgb]{1.00,0.00,0.00}{\textbf{0.0447}} \\ \hline
		
	\end{tabular}
\end{table*}

\begin{table*}[t]
	\renewcommand\arraystretch{1.2} 
	\caption{SAD comparisons on the Jasper Ridge dataset, where the top two values are marked as \textcolor[rgb]{1.00,0.00,0.00}{\textbf{red}} and \textcolor[rgb]{0.00,0.00,1.00}{\textbf{blue}}.}\label{jaspersad}
		\vspace{-0.15cm}
	\centering
	\setlength{\tabcolsep}{1mm} 
	\begin{tabular}{|c|c|c|c|c|c|c|c|c|c|c|}
		\hline
		~ {\textbf{Cases}} ~~&~ \textbf{VCA-FCLS} ~ &~  {\textbf{MVC-NMF}}~  &~ {\textbf{SNMF}} ~ & ~ {\textbf{RNMF}} ~  &~ {\textbf{SGSNMF}} ~& ~  {\textbf{IRNMF}} ~ &~ {\textbf{HGNMF-FS}}~ & ~ {\textbf{MV-DUAL}} ~& ~ {\textbf{A2SAN}} ~&~{\textbf{MOGNMF}}~ \\ 
		\hline \hline
		Tree& 0.1369&	0.1149&	0.1370&	0.1441&	0.0970&	0.0967&	\textcolor[rgb]{0.00,0.00,1.00}{\textbf{0.0810}}&	0.1068&0.0990 & \textcolor[rgb]{1.00,0.00,0.00}{\textbf{0.0776}} \\ \hline
		Soil&0.1367&	0.1826&	0.1215&	0.1328&	0.1572&	0.1554&	\textcolor[rgb]{0.00,0.00,1.00}{\textbf{0.1187}}&	0.3781&0.2051 & \textcolor[rgb]{1.00,0.00,0.00}{\textbf{0.1147}} \\ \hline
		Water&0.1066&	0.0617&	0.0932&	0.1052&	0.0455&	0.0454&	\textcolor[rgb]{0.00,0.00,1.00}{\textbf{0.0543}}&	0.0407&0.0914& \textcolor[rgb]{1.00,0.00,0.00}{\textbf{0.0374}} \\ \hline
		Road&0.0559&	0.0303&	0.0754&	0.0664&	0.0315&	\textcolor[rgb]{0.00,0.00,1.00}{\textbf{0.0283}}&	0.0335&	0.0427&0.0935
		& \textcolor[rgb]{1.00,0.00,0.00}{\textbf{0.0279}} \\ \hline \hline
		Average & 0.1090&	0.0974&	0.1068&	0.1121&	0.0828 &0.0814&	\textcolor[rgb]{0.00,0.00,1.00}{\textbf{0.0696}}&	0.1421 &0.1223 & \textcolor[rgb]{1.00,0.00,0.00}{\textbf{0.0644}} \\ \hline
	\end{tabular}
\end{table*}

\begin{table*}[t]
	\renewcommand\arraystretch{1.2} 
	\caption{SAD comparisons on the Urban dataset, where the top two values are marked as \textcolor[rgb]{1.00,0.00,0.00}{\textbf{red}} and \textcolor[rgb]{0.00,0.00,1.00}{\textbf{blue}}.}\label{urbansad}
		\vspace{-0.15cm}
	\centering
	\setlength{\tabcolsep}{1mm} 
	\begin{tabular}{|c|c|c|c|c|c|c|c|c|c|c|}
		\hline
	~ {\textbf{Cases}} ~~&~ \textbf{VCA-FCLS} ~ &~  {\textbf{MVC-NMF}}~  &~ {\textbf{SNMF}} ~ & ~ {\textbf{RNMF}} ~  &~ {\textbf{SGSNMF}} ~& ~  {\textbf{IRNMF}} ~ &~ {\textbf{HGNMF-FS}}~ & ~ {\textbf{MV-DUAL}} ~& ~ {\textbf{A2SAN}} ~&~{\textbf{MOGNMF}}~ \\ 
		\hline \hline
		Asphalt & 0.1358&	0.1490&	0.1362&	0.1243&0.1152&	0.1395&	0.1282&	0.1699&\textcolor[rgb]{1.00,0.00,0.00}{\textbf{0.0969}}&		\textcolor[rgb]{0.00,0.00,1.00}{\textbf{0.1097 }}\\ \hline
		Grass& 0.1159&	0.1083&	0.1292&	0.1361&	0.1266&	0.1125&	0.1149&	0.1281&\textcolor[rgb]{1.00,0.00,0.00}{\textbf{0.0951}}&	\textcolor[rgb]{0.00,0.00,1.00}{\textbf{0.1051}} \\ \hline
		Tree & 0.0562&	0.0837&	0.0880&	0.0891&	\textcolor[rgb]{0.00,0.00,1.00}{\textbf{0.0481}}&	0.0814&	0.0681&	0.1248&\textcolor[rgb]{1.00,0.00,0.00}{\textbf{0.0477}} &	0.0647\\ \hline
		Roof & 0.1021&	0.1255&	0.0857&	0.1005&		\textcolor[rgb]{0.00,0.00,1.00}{\textbf{0.0686}} &0.0691&	0.0547&	0.1057&0.1290&	\textcolor[rgb]{1.00,0.00,0.00}{\textbf{0.0595}}\\ \hline \hline
		Average & 0.1025&	0.1166&	0.1098&	0.1125&	\textcolor[rgb]{0.00,0.00,1.00}{\textbf{0.0896}}&	0.0974&	0.0915&	0.1321&	0.0919&\textcolor[rgb]{1.00,0.00,0.00}{\textbf{0.0848}} \\ \hline
	\end{tabular}
\end{table*}

\begin{table*}[!ht]
	\renewcommand\arraystretch{1.2} 
	\caption{{SAD values on the Samon dataset under different SNR conditions, where the top two values are marked as \textcolor[rgb]{1.00,0.00,0.00}{\textbf{red}} and \textcolor[rgb]{0.00,0.00,1.00}{\textbf{blue}}. }}\label{SAMONTABLE}
		\vspace{-0.15cm}
	\centering
	\setlength{\tabcolsep}{1mm}  
	\begin{tabular}{|c|c|c|c|c|c|c|c|c|c|c|}
		\hline
	~ {\textbf{Cases}} ~~&~ \textbf{VCA-FCLS} ~ &~  {\textbf{MVC-NMF}}~  &~ {\textbf{SNMF}} ~ & ~ {\textbf{RNMF}} ~  &~ {\textbf{SGSNMF}} ~& ~  {\textbf{IRNMF}} ~ &~ {\textbf{HGNMF-FS}}~ & ~ {\textbf{MV-DUAL}} ~& ~ {\textbf{A2SAN}} ~&~{\textbf{MOGNMF}}~ \\ 
		\hline \hline
		SNR=10 & 0.1338	&0.1278	&0.1221&	0.1127&	0.1361	&0.1041	&	 \textcolor[rgb]{0.00,0.00,1.00}{\textbf{0.1007}}	&	0.1305	& 0.1209	&\textcolor[rgb]{1.00,0.00,0.00}{\textbf{0.0964}} \\ \hline
		SNR=20  & 0.0648&	0.0851	&0.0636&	0.0907&	0.0938&	\textcolor[rgb]{1.00,0.00,0.00}{\textbf{0.0595}}	&	0.0601	&	0.0969	&0.0864	& \textcolor[rgb]{0.00,0.00,1.00}{\textbf{0.0599}}  \\ \hline
	SNR=30& 0.0603&	0.0620&	0.0563	&0.0608	&0.0619	&\textcolor[rgb]{1.00,0.00,0.00}{\textbf{0.0492}}	&	0.0557	&	0.0750	&0.0597&	 \textcolor[rgb]{0.00,0.00,1.00}{\textbf{0.0530}}  \\ \hline
	SNR=40 & 0.0596	&0.0629&	0.0761&	0.0521	&0.0557&	0.0502	&	 \textcolor[rgb]{0.00,0.00,1.00}{\textbf{0.0497}}	&	0.0501&	0.0631&	\textcolor[rgb]{1.00,0.00,0.00}{\textbf{0.0464}}  \\ \hline\hline
		Average &0.0721&	0.0901&	0.0795&	0.0891&	0.0869&	\textcolor[rgb]{0.00,0.00,1.00}{\textbf{0.0658}}	&	0.0666&		0.0881&	0.0825& \textcolor[rgb]{1.00,0.00,0.00}{\textbf{0.0639}}\\ \hline
	\end{tabular}
\end{table*}

\begin{figure}[t]
	\centering
	\subfigcapskip=-3pt
	\subfigure[]{
		\centering
		\includegraphics[width=0.15\textwidth]{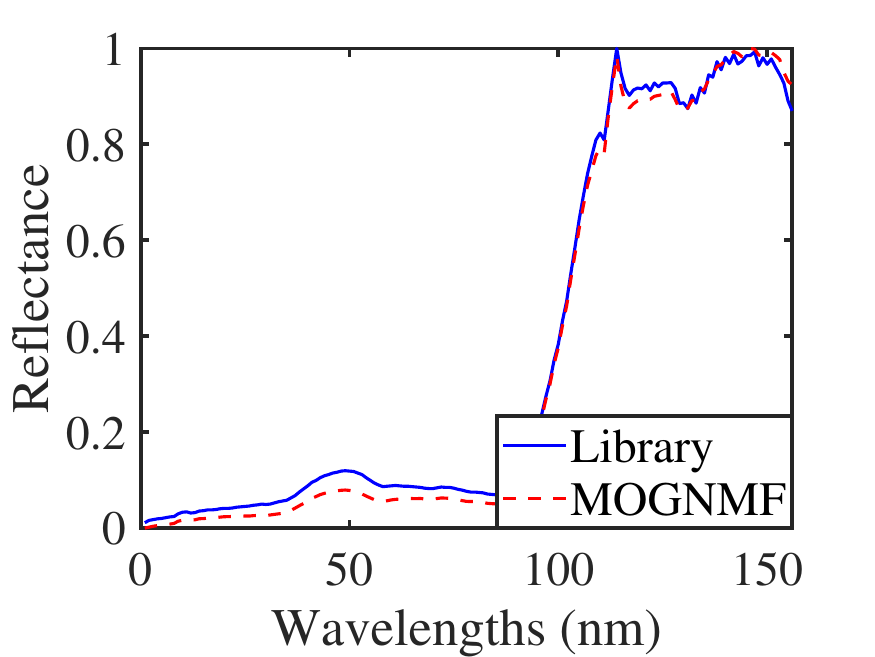}  
	}
	\hspace{-0.4cm}
	\subfigcapskip=-3pt
	\subfigure[]{
		\centering
		\includegraphics[width=0.15\textwidth]{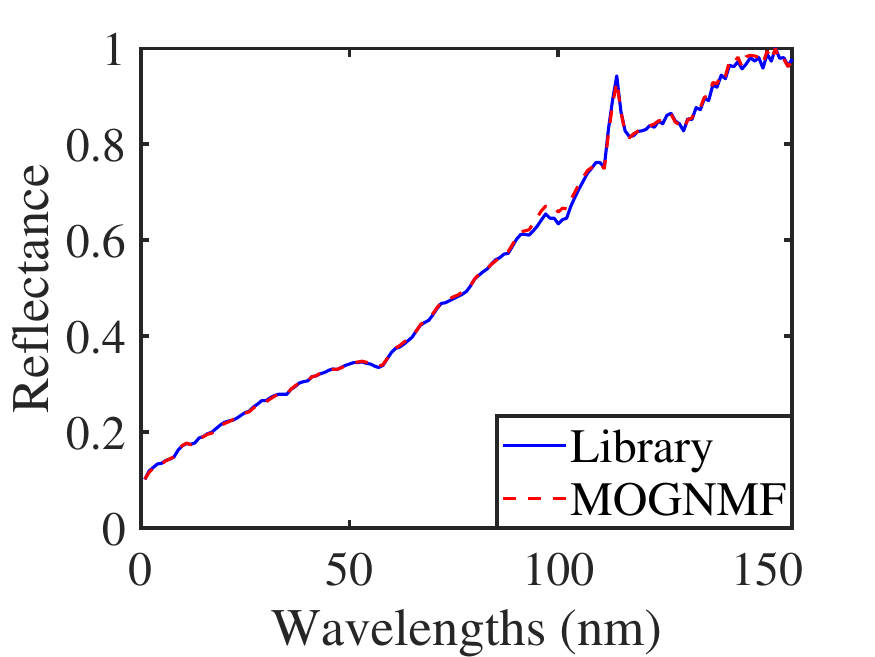}  
	}
	\hspace{-0.4cm}
	\subfigcapskip=-3pt
	\subfigure[]{
		\centering
		\includegraphics[width=0.15\textwidth]{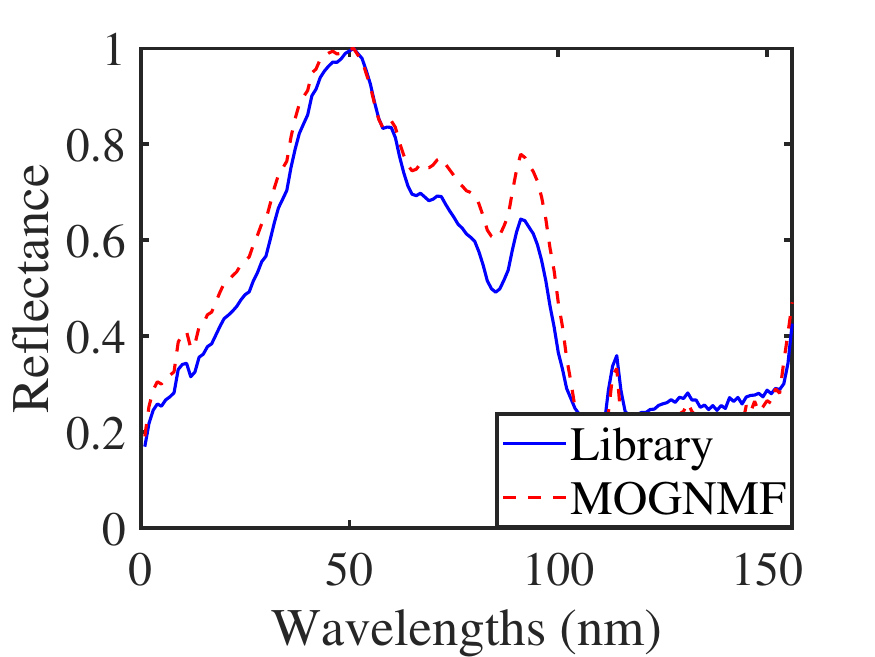} 
	}
	\vspace{-0.2cm}
	\caption{Endmember signatures extracted by MOGNMF on the Samson dataset, where (a) Soil, (b) Tree, and (c) Water.} \label{RELF1}
\end{figure}

\begin{figure}[t]
	\centering
	\subfigcapskip=-3pt
	\subfigure[]{
		\centering
		\includegraphics[width=0.18\textwidth]{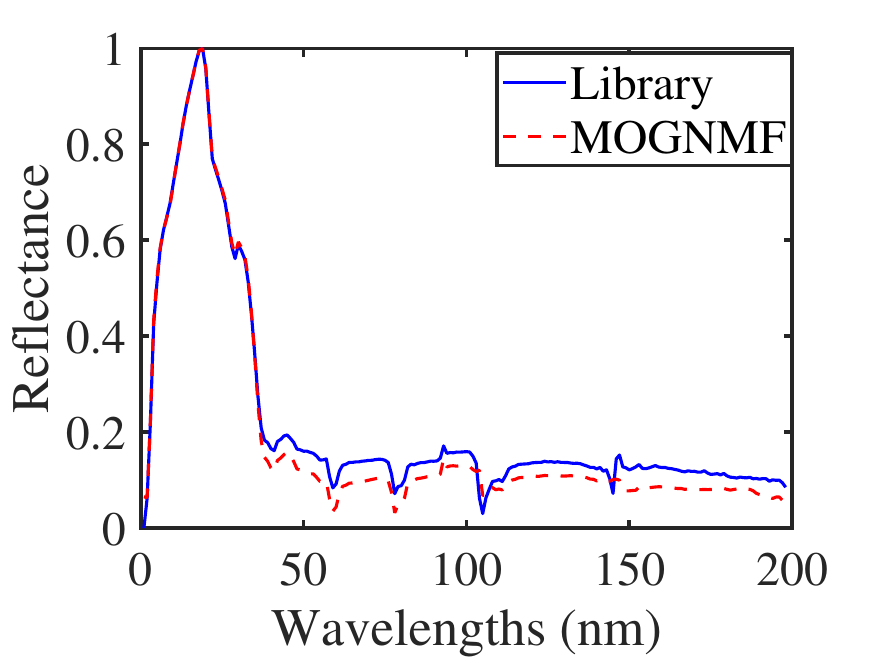}
	}
	\subfigcapskip=-3pt
	\subfigure[]{
		\centering
		\includegraphics[width=0.18\textwidth]{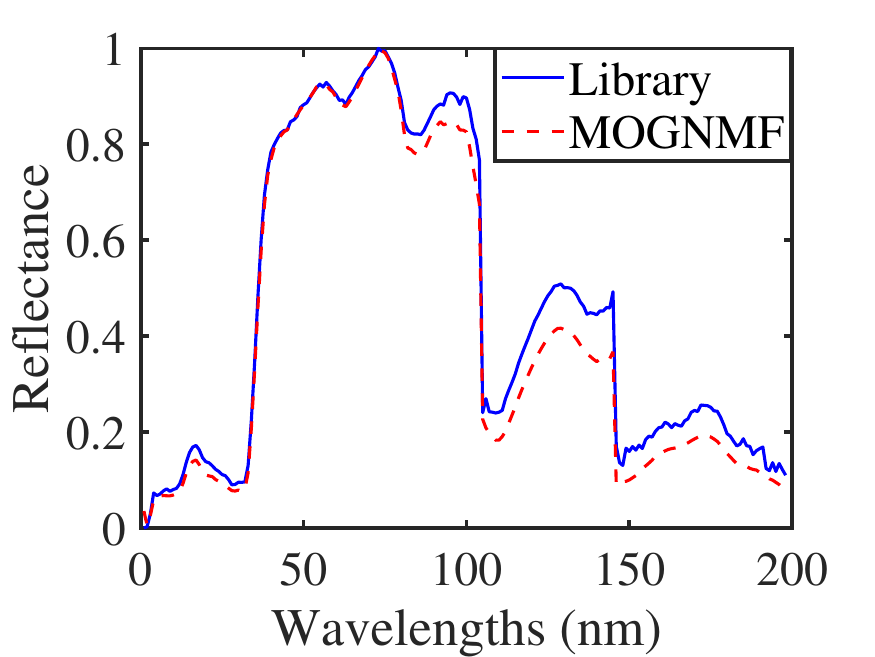}
	}
	\\ 
	\subfigcapskip=-3pt
	\subfigure[]{
		\centering
		\includegraphics[width=0.18\textwidth]{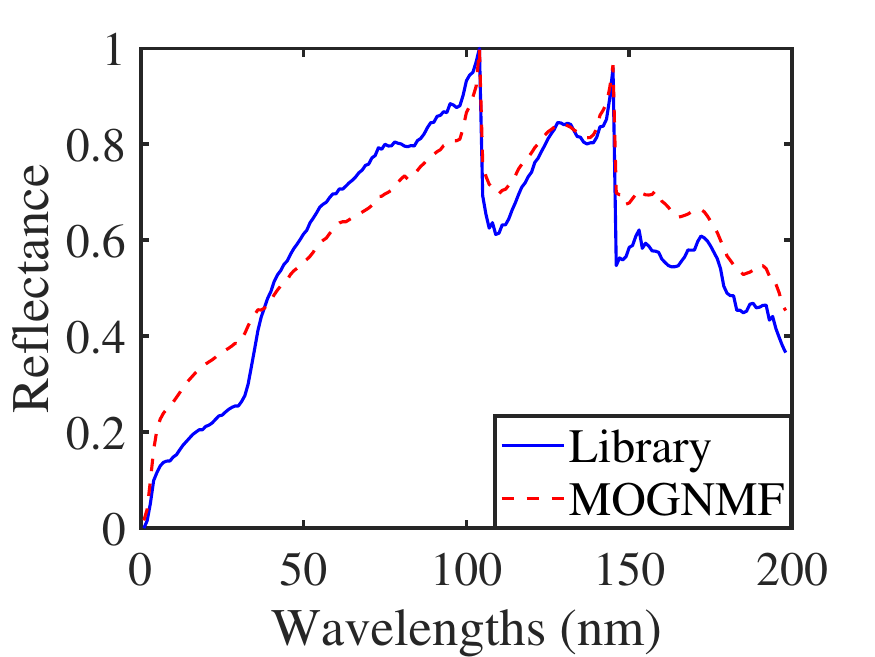}
	}
	\subfigcapskip=-3pt
	\subfigure[]{
		\centering
		\includegraphics[width=0.18\textwidth]{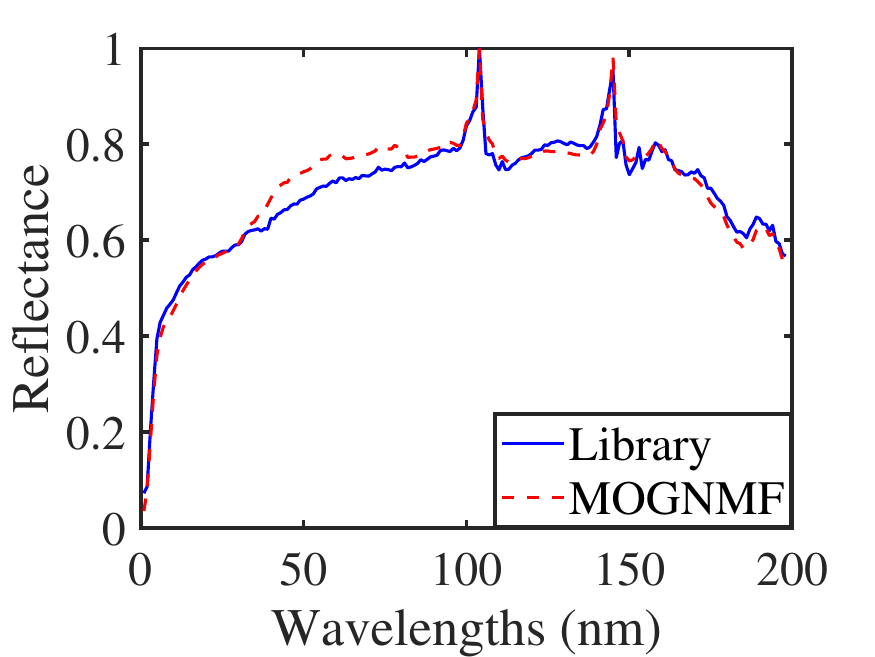}
	}
	\vspace{-0.2cm}
	\caption{Endmember signatures extracted by MOGNMF on the Jasper Ridge dataset, where (a) Tree, (b) Soil, (c) Water, and (d) Road.}  \label{RELF2}
\end{figure}
\begin{figure}[t]
	\centering
	\subfigcapskip=-3pt
	\subfigure[]{
		\centering
		\includegraphics[width=0.18\textwidth]{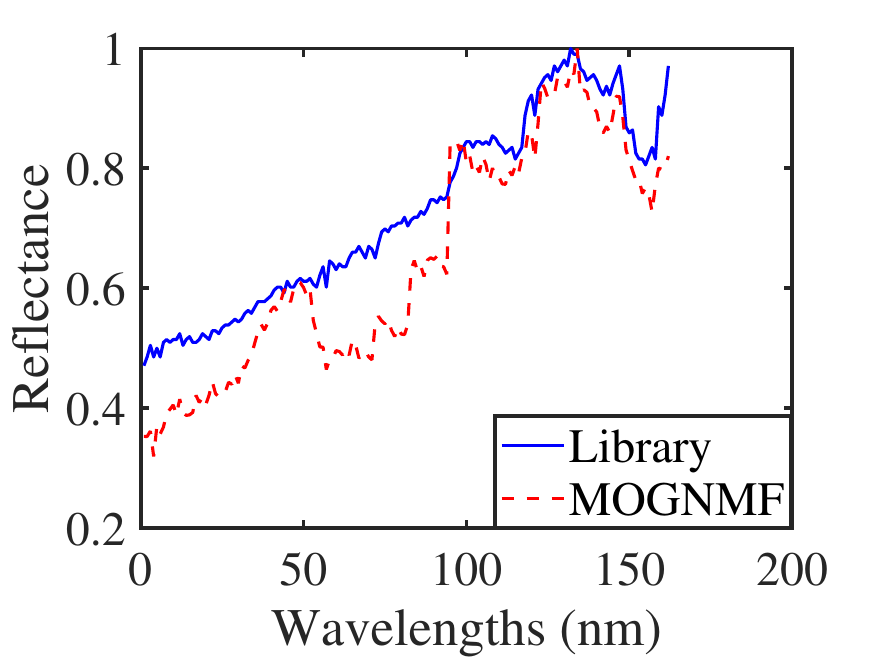}
	}
	\subfigcapskip=-3pt
	\subfigure[]{
		\centering
		\includegraphics[width=0.18\textwidth]{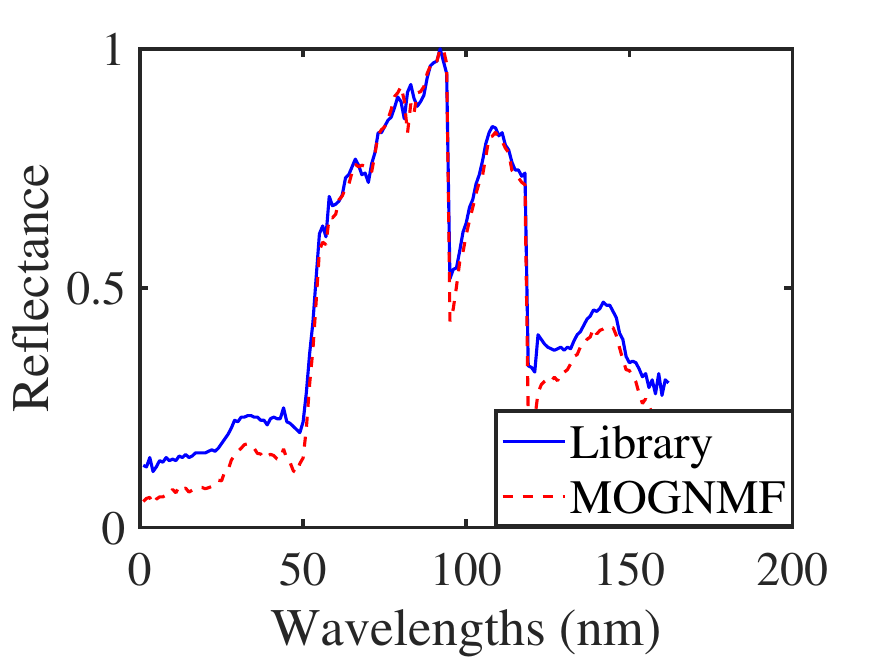}
	}
	\\ 
	\subfigcapskip=-3pt
	\subfigure[]{
		\centering
		\includegraphics[width=0.18\textwidth]{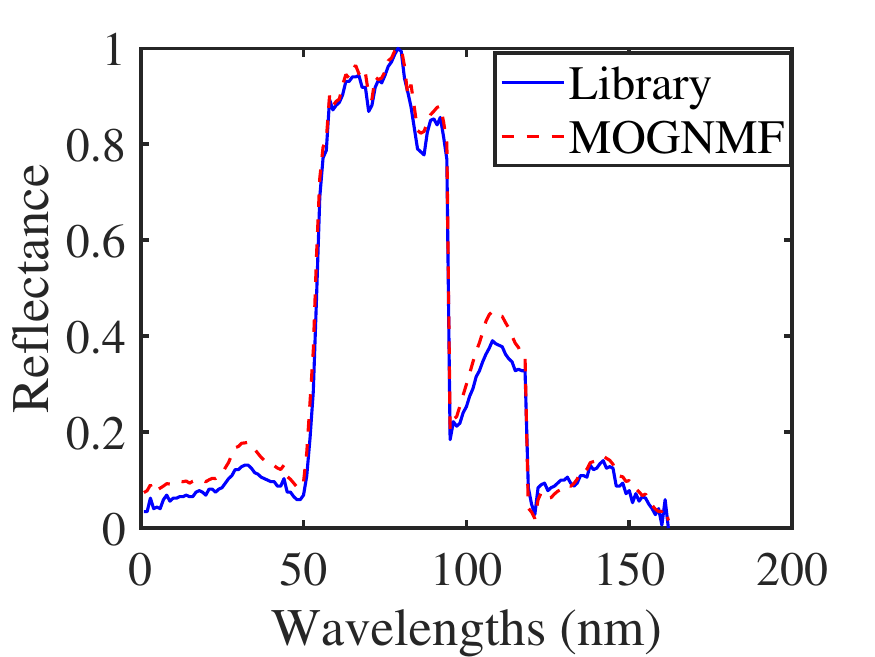}
	}
	\subfigcapskip=-3pt
	\subfigure[]{
		\centering
		\includegraphics[width=0.18\textwidth]{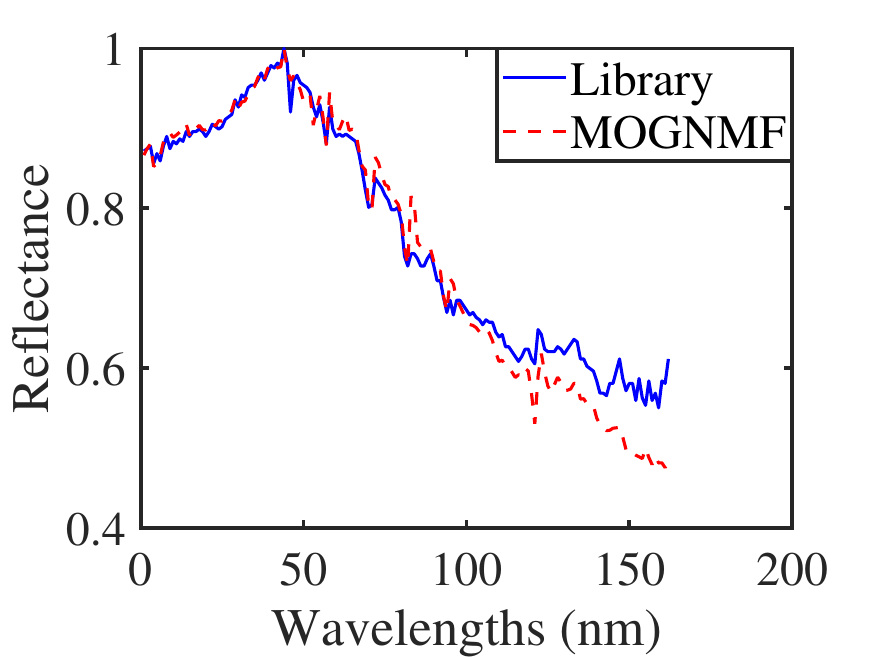}
	}
	\vspace{-0.2cm}
	\caption{Endmember signatures extracted by MOGNMF on the Urban dataset, where (a) Asphalt, (b) Grass, (c) Tree, and (d) Roof.}  \label{urban}
\end{figure}

\begin{figure*}[!ht]

    \begin{minipage}{\textwidth}
        \centering
         \vspace{-1cm}
        \includegraphics[width=0.95\textwidth]{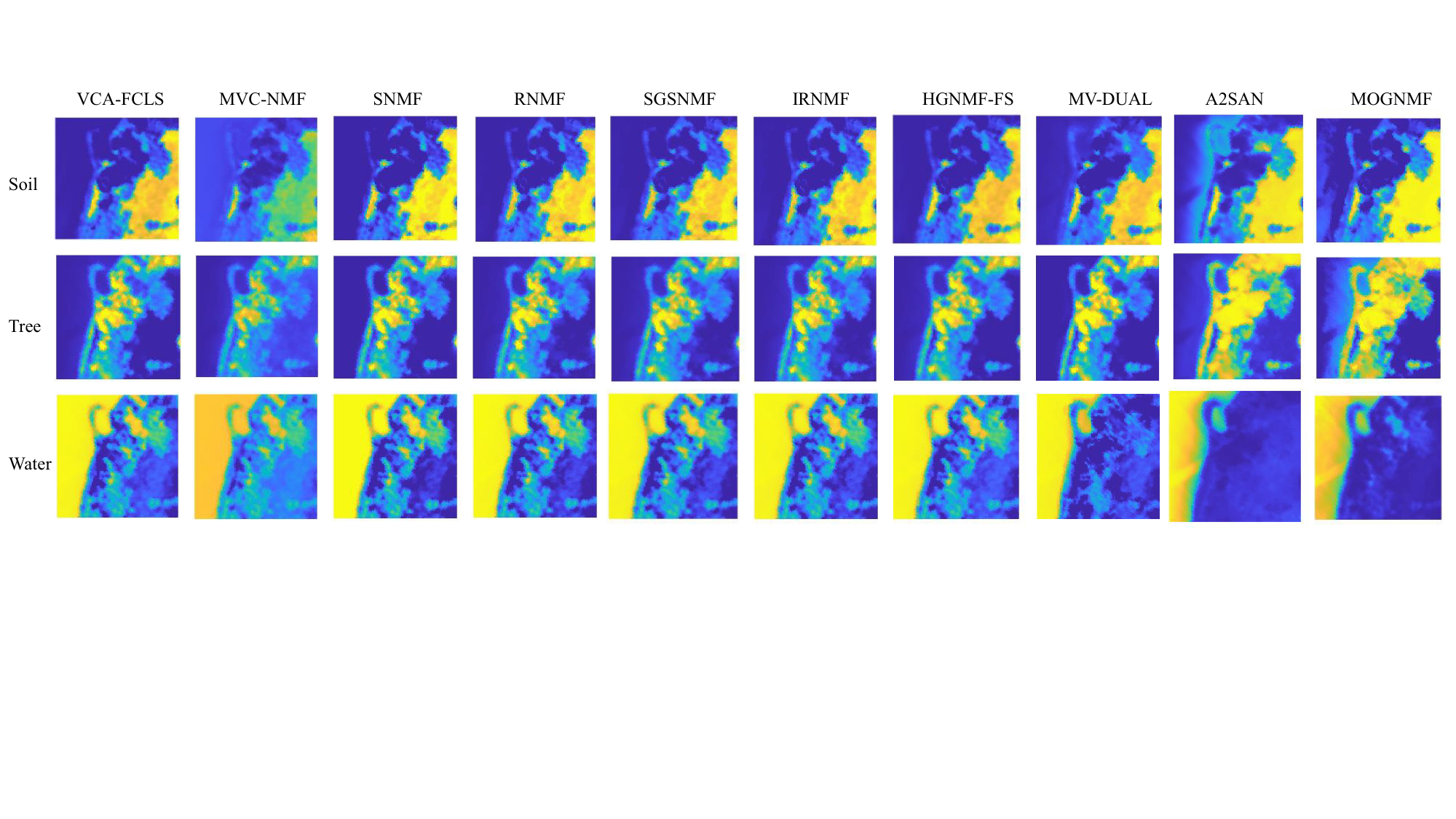}
        \vspace{-3.8cm}
        \caption{Abundance map comparisons on the Samson  dataset.}
        \label{samonrmse}
    \end{minipage}
    
    \vspace{0.5cm}

    \begin{minipage}{\textwidth}
        \centering
         \vspace{-1cm}
        \includegraphics[width=0.95\textwidth]{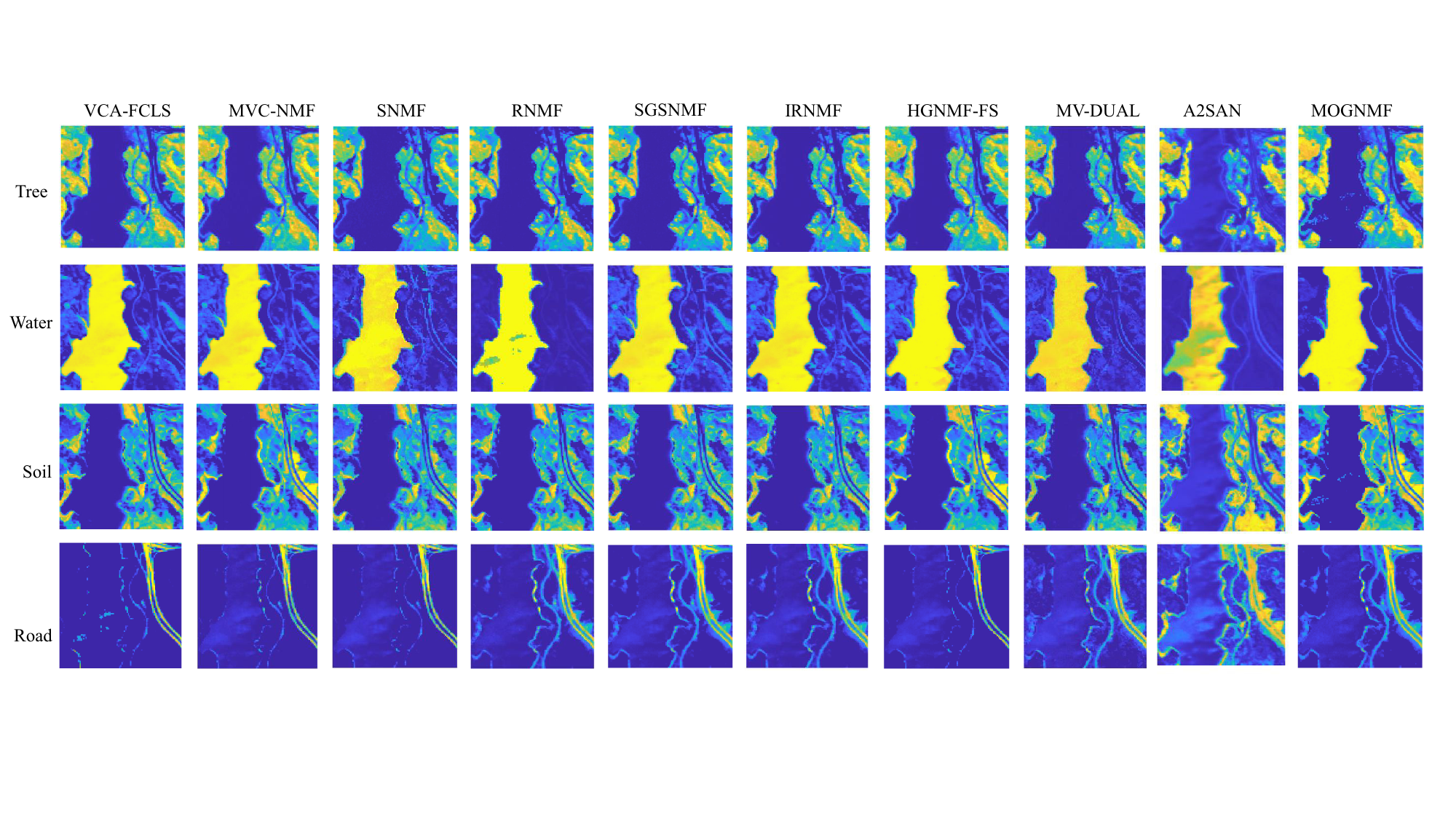}
        \vspace{-2cm}
        \caption{Abundance map comparisons on the Jasper Ridge dataset.}
        \label{jasperrmse}
    \end{minipage}
       \vspace{0.5cm}

    \begin{minipage}{\textwidth}
        \centering
         \vspace{-1cm}
        \includegraphics[width=0.95\textwidth]{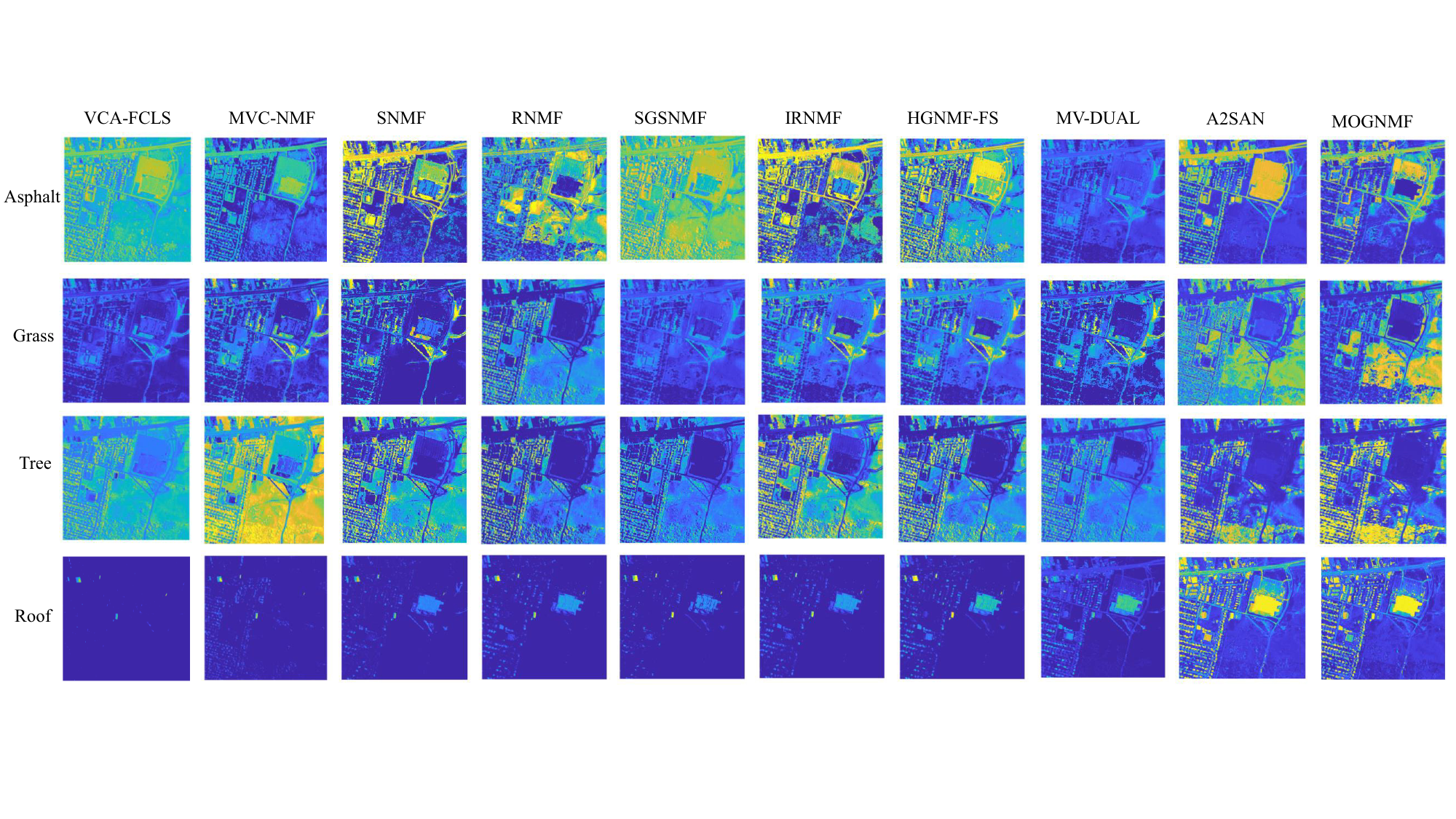}
        \vspace{-1.9cm}
        \caption{Abundance map comparisons on the Urban dataset.}
        \label{urbanrmse}
    \end{minipage}
\end{figure*}

Fig. \ref{DC2EM} presents the abundance maps obtained by different unmixing methods on the Simu-2 dataset with \( \text{SNR}=30 \, \text{dB} \), where EM1-EM4 denote the selected endmembers. In addition, Fig. \ref{DC2SNR} shows the abundance maps of the endmember 1 under different noise levels. It can be concluded that, under different SNR conditions, the proposed MOGNMF can generate smoother abundance maps compared with other methods. Even in the case of high noise, the proposed method can still accurately estimate the land cover distribution, which shows its robustness to noise.

\subsection{Results on Real Datasets}\label{Results on Real-world Datasets}

\subsubsection{Dataset Description} 

The Samson dataset\footnote{\url{http://www.escience.cn/people/feiyunZHU/Dataset_GT.html}} is extensively used in unmixing, which contains $952 \times 952$ pixels, each with 156 spectral channels, covering the wavelength range from 401 nm to 889 nm, and the spectral resolution is up to 3.13 nm. A subregion of $95 \times 95$ pixels starting at position $(252, 332)$ is selected for analysis. Specifically, this image includes three targets, i.e., Soil, Tree, and Water.

The Jasper Ridge dataset\footnote{\url{http://www.escience.cn/system/file?fileId=68574}} consists of $512 \times 614$ pixels, with each pixel recorded across 224 spectral channels covering a wavelength range from 380 nm to 2500 nm, and a spectral resolution of up to 9.46 nm. A subimage of $100 \times 100$ pixels is used for analysis. The endmembers include four categories: Tree,  Soil,  Water, and Road.

The Urban dataset consists\footnote{\url{https://rslab.ut.ac.ir/data}} of $307 \times 307$ pixels, with each pixel recorded across 210 spectral bands covering a wavelength range from 400 nm to 2500 nm. After removing bands 1–4, 76, 87, 101–111, 136–153, and 198–210 due to dense water vapor absorption and atmospheric effects, 162 bands are retained for analysis. The dataset contains four endmembers: Asphalt, Grass, Tree, and Roof.

\begin{figure}[t]
	\setlength{\subfigcapskip}{-5pt}  
	\subfigure[ ]{
		\includegraphics[scale=0.26]{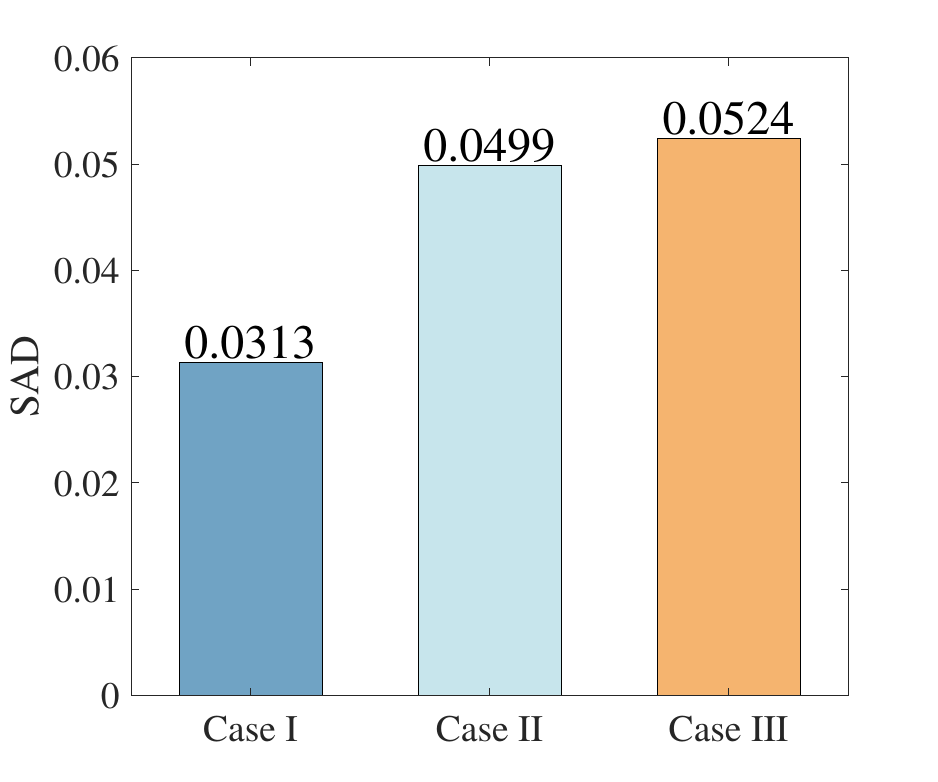} 
	}
	\subfigure[ ]{			
		\includegraphics[scale=0.26]{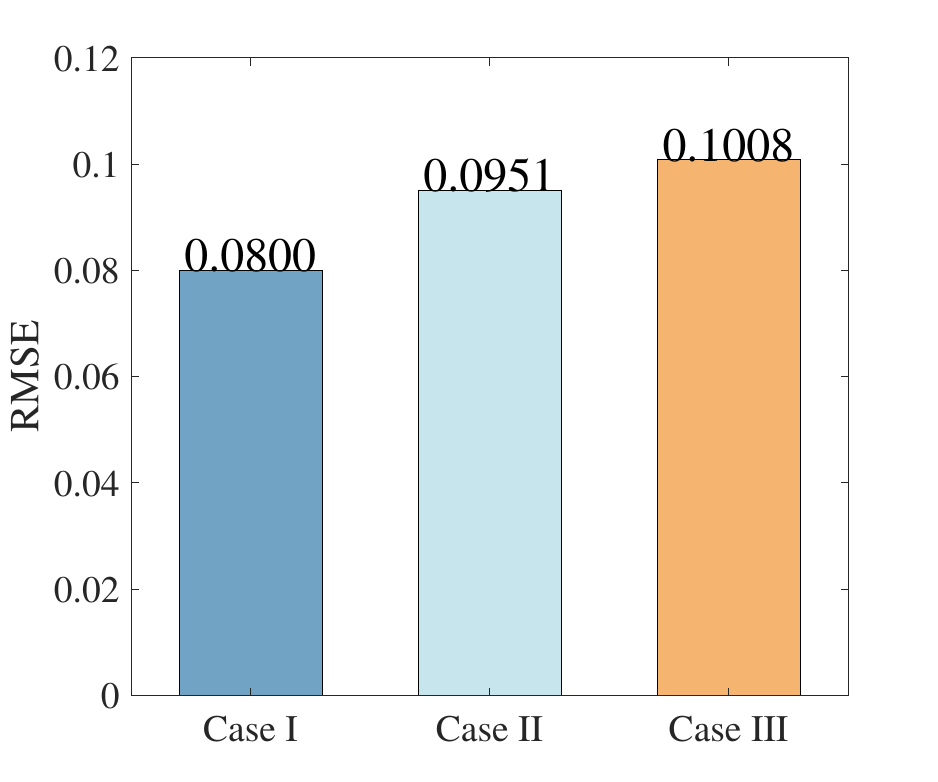}
	}
	\vspace{-0.4cm}
	\caption{Ablation comparisons of regularization terms on the Simu-1 dataset, where (a) SAD and (b) RMSE.}\label{fig:ablation1}
\end{figure}

\begin{figure}[t]
    \setlength{\subfigcapskip}{-5pt} 
    \centering
    \subfigure[ ]{
        \includegraphics[scale=0.259]{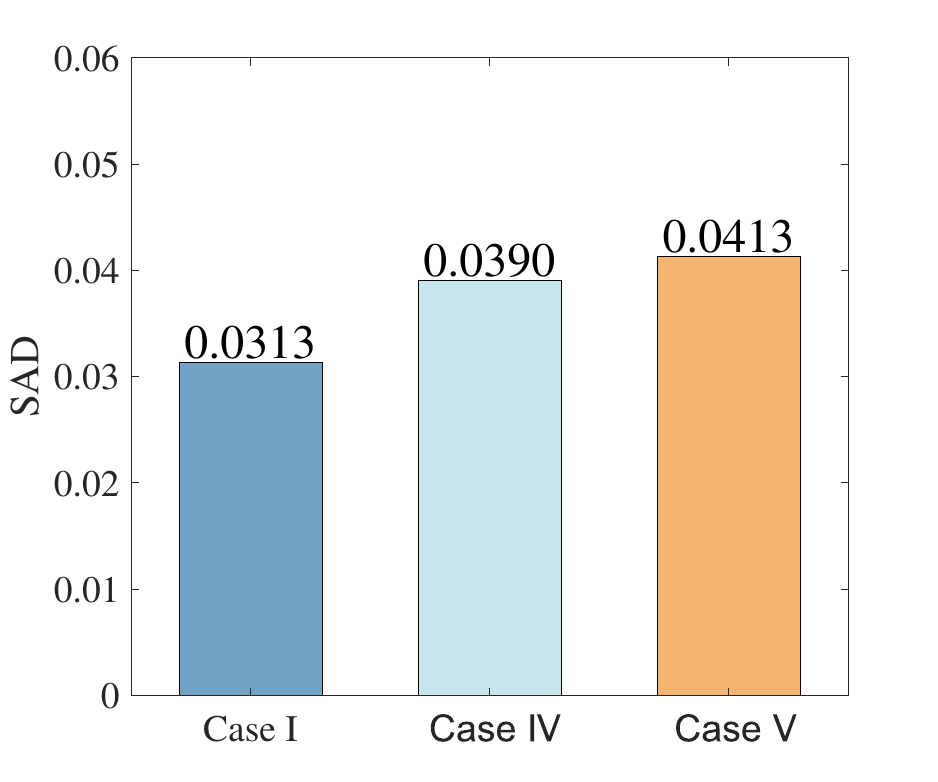}
    }
    \subfigure[ ]{
        \includegraphics[scale=0.259]{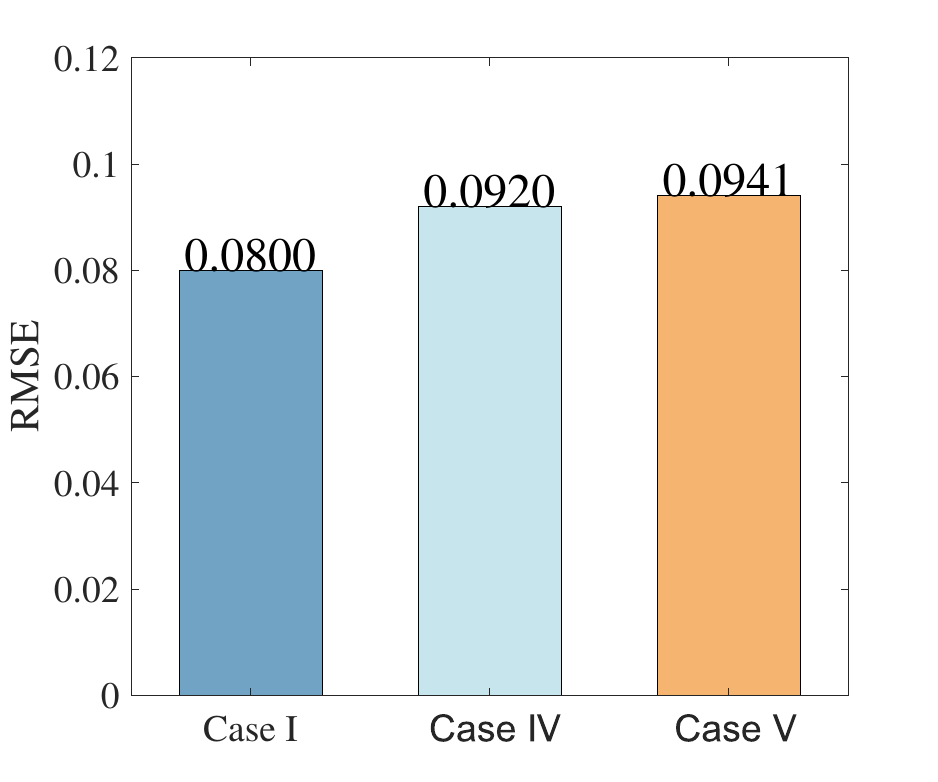}
    }
    \vspace{-0.4cm}
    \caption{Ablation comparisons of graph regularization orders on the Simu-1 dataset, where (a) SAD and (b) RMSE.}
    \label{fig:ablation2}
\end{figure}

\subsubsection{Numerical Results}

To demonstrate the performance of the proposed MOGNMF on real-world hyperspectral datasets, the SAD between the estimated endmembers and the reference spectra from the USGS spectral library is computed. The SAD values for MOGNMF and other compared methods on the Samson dataset are summarized in Table \ref{samonsad}. The results show that MOGNMF achieves better performance than competitors. To illustrate the capability of MOGNMF in extracting endmember signatures, Fig. \ref{RELF1} presents a comparison between the extracted signatures and the reference endmembers. The high degree of consistency between them validates the effectiveness of MOGNMF in the HU task.
Besides, Table \ref{jaspersad} shows the resulting SAD obtained by different methods on the Jasper Ridge  dataset. It can be observed that our proposed MOGNMF also achieves a lower average SAD value than others. Fig. \ref{RELF2} presents a comparison between the endmember signatures extracted by MOGNMF and the reference endmember signatures, demonstrating that MOGNMF can accurately extract spectral features and effectively distinguish different endmember components. Furthermore, to further validate the unmixing capability of MOGNMF on a larger real dataset, the Urban dataset is used. Table \ref{urbansad} presents the SAD values of different methods on the Urban dataset. The results demonstrate that MOGNMF also performs well on the Urban dataset, and the extracted endmember signatures by MOGNMF, compared with the reference spectra, are shown in Fig. \ref{urban}, confirming the effectiveness of the proposed method.
 
Additionally, Fig. \ref{samonrmse} presents the abundance maps estimated by various methods on the Samson dataset, where blue represents lower abundance values, and yellow represents higher abundance values, which closely match the overall distribution trend of the reference map. The proposed MOGNMF demonstrates significant advantages in preserving spatial details and more naturally handling boundary transitions. For the Soil components, the proposed MOGNMF excels in retaining the edge details of the Soil regions, providing smoother transitions, and avoiding excessive sparsity or over-concentration in certain areas. In contrast, VCA-FCLS and MVC-NMF show blurring at the boundaries of Soil regions, with noticeable regional errors in abundance distribution, while the abundance maps produced by IRNMF and SGSNMF show both overestimated and underestimated abundance values. These observations highlight the advantages of the multi-order graph regularization in the proposed MOGNMF. Moreover, Fig. \ref{jasperrmse} shows the abundance maps estimated by various methods on the Jasper Ridge  dataset. The abundance maps obtained by our proposed MOGNMF are highly consistent with the true distribution, effectively recovering the true abundance distribution of the hyperspectral data. Similarly, Fig. \ref{urbanrmse} presents the abundance maps, indicating that MOGNMF also produces abundance results on the Urban dataset that are highly consistent with the ground truth.

\subsection{Ablation Studies}\label{Ablation Studies}

In this subsection, ablation experiments are conducted to verify the effectiveness of the multi-order graph and sparsity in our proposed MOGNMF. We first analyze the impact by removing different regularization terms as follows.
\begin{itemize}
  \item Case I: The proposed MOGNMF (also \eqref{main})
    \begin{equation}
  	\begin{aligned}
  		\min_{\mathbf{A}, \mathbf{S}, \mathbf{E} } \quad &\frac{1}{2} \| \mathbf{X}-\mathbf{E}  - \mathbf{A} \mathbf{S} \|_F^2 + \gamma \| \mathbf{S} \|_{1/2} \\
  		& \quad + \beta \| \mathbf{E} \|_{2,1} + \frac{\lambda}{2} \text{Tr}(\mathbf{S} \mathbf{L}_m \mathbf{S}^\top).  
  	\end{aligned}
  \end{equation}
  \item Case II: The proposed MOGNMF \textit{without} the noise matrix term 
  \begin{equation}
  	\begin{aligned}
  		\min_{\mathbf{A}, \mathbf{S}} ~ \frac{1}{2} \| \mathbf{X} - \mathbf{A} \mathbf{S} \|_F^2 + \gamma \| \mathbf{S} \|_{1/2} + \frac{\lambda}{2} \text{Tr}(\mathbf{S} \mathbf{L}_m \mathbf{S}^\top).
  	\end{aligned}
  \end{equation}
  \item Case III: The proposed MOGNMF  \textit{without} the noise matrix term and graph regularization term
\begin{equation}
	\begin{aligned}
		\min_{\mathbf{A}, \mathbf{S} }~\frac{1}{2} \| \mathbf{X}- \mathbf{A} \mathbf{S} \|_F^2 + \gamma \| \mathbf{S} \|_{1/2}.
	\end{aligned}
\end{equation}
\end{itemize}
  Fig. \ref{fig:ablation1} presents the ablation results on the Simu-1 dataset with \( \text{SNR}=20 \, \text{dB} \). It can be observed that, compared with the two simplified models in Case II and Case III, our proposed MOGNMF in Case I achieves better performance in the unmixing task, demonstrating the importance and effectiveness of combining all three regularization terms.
  
  Next, we analyze the unmixing results when only using the first-order and second-order nearest neighbor relationships by comparing Case I and the following two models.
\begin{itemize}
  \item Case IV: The proposed MOGNMF  \textit{with}  an only second-order graph regularization term
    \begin{equation}
  	\begin{aligned}
  		\min_{\mathbf{A}, \mathbf{S}, \mathbf{E} } \quad &\frac{1}{2} \| \mathbf{X}-\mathbf{E}  - \mathbf{A} \mathbf{S} \|_F^2 + \gamma \| \mathbf{S} \|_{1/2} \\
  		& \quad + \beta \| \mathbf{E} \|_{2,1} + \frac{\lambda}{2} \text{Tr}(\mathbf{S} \mathbf{L}_2 \mathbf{S}^\top).
  	\end{aligned}
  \end{equation}
  \item Case V: The proposed MOGNMF \textit{with}   an only first-order graph regularization term
    \begin{equation}
  	\begin{aligned}
  		\min_{\mathbf{A}, \mathbf{S}, \mathbf{E} } \quad &\frac{1}{2} \| \mathbf{X}-\mathbf{E}  - \mathbf{A} \mathbf{S} \|_F^2 + \gamma \| \mathbf{S} \|_{1/2} \\
  		& \quad + \beta \| \mathbf{E} \|_{2,1} + \frac{\lambda}{2} \text{Tr}(\mathbf{S} \mathbf{L}_1 \mathbf{S}^\top).
  	\end{aligned}
  \end{equation}
\end{itemize}
From Fig. \ref{fig:ablation2}, it is found that our proposed MOGNMF in Case I can still obtain better unmixing performance than Case IV and Case V, which highlights the superiority of the adaptive multi-order graph framework.
\begin{figure}[!ht]
    \centering
    \setlength{\subfigcapskip}{-2pt}  

    \subfigure[ ]{
        \includegraphics[scale=0.259]{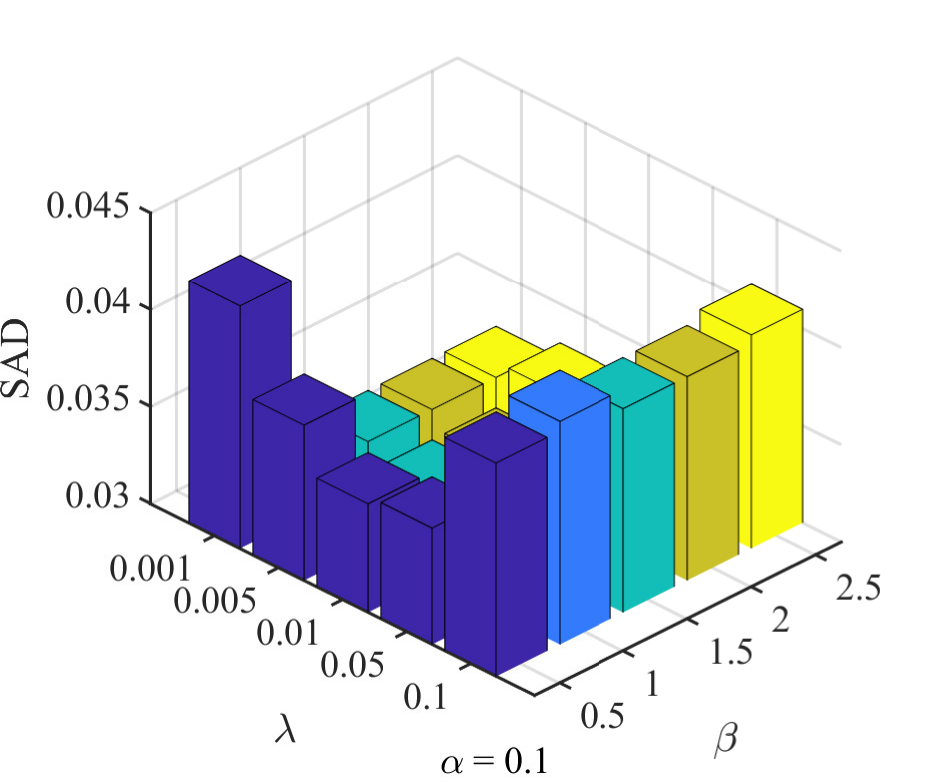}
    }
    \subfigure[ ]{
        \includegraphics[scale=0.259]{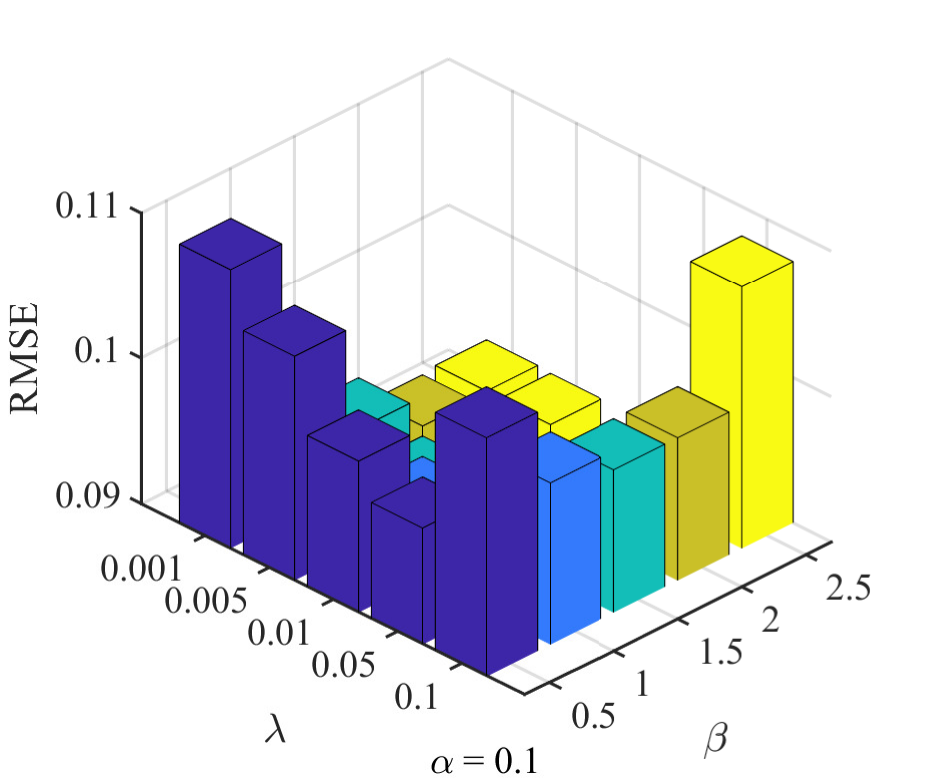}
    }
    \vspace{1ex}
    \subfigure[ ]{
        \includegraphics[scale=0.259]{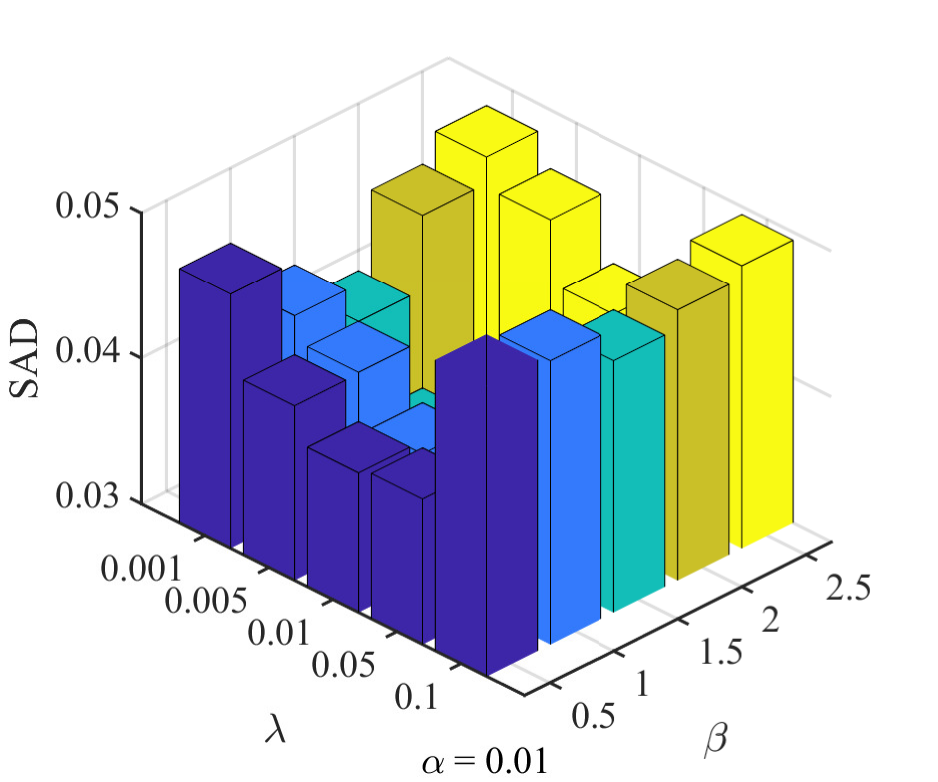}
    }
    \subfigure[ ]{
        \includegraphics[scale=0.259]{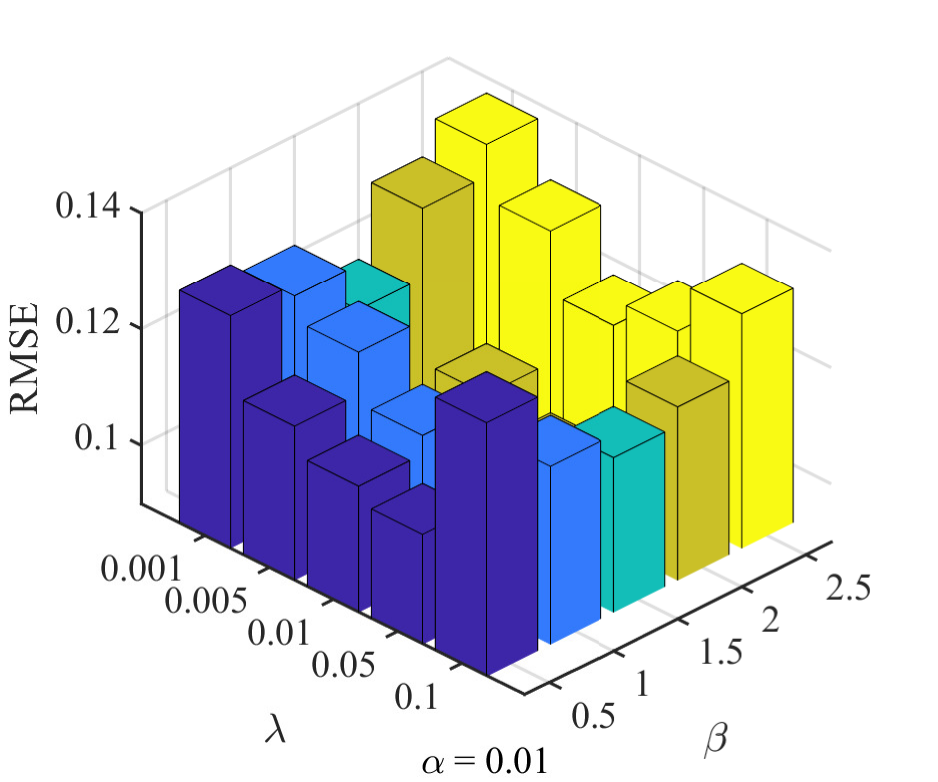}
    }
    \vspace{-0.45cm}
    \caption{Parameter sensitivity analysis with respect to $\lambda$ and $\beta$, where (a)-(b) are results when \(\alpha = 0.1\),  (c)-(d) are results when \(\alpha = 0.01\).}
    \label{PARAMETER}
\end{figure}

\subsection{Discussion}\label{Discussion}

This section discusses the parameter analysis, impact of image sizes, impact of endmember numbers, impact of order \( K \), convergence analysis, sensitivity analysis, robustness analysis, and convergence analysis.

\subsubsection{Parameter Analysis}

Four critical parameters, namely \( \alpha, \beta, \gamma, \lambda \), are considered in this study, where \(\gamma\) is estimated according to \eqref{gamma}. Therefore, we first analyze the influence of \(\beta\) and \(\lambda\) while fixing \(\alpha = 0.1, 0.01\). As shown in Fig.~\ref{PARAMETER}, under the condition of SNR = 20 and six endmembers, SAD and RMSE exhibit similar variation trends. Specifically, both metrics reach their minimum values when \(\lambda\) is within the range of 0.01 to 0.1 and \(\beta\) is within the range of 1 to 2. Hence, our experiments are conducted using parameter values within these ranges.

\subsubsection{Impact of Image Sizes} 

\begin{figure}[t]
	\setlength{\subfigcapskip}{-2pt}  
	\subfigure[ ]{
		\includegraphics[scale=0.26]{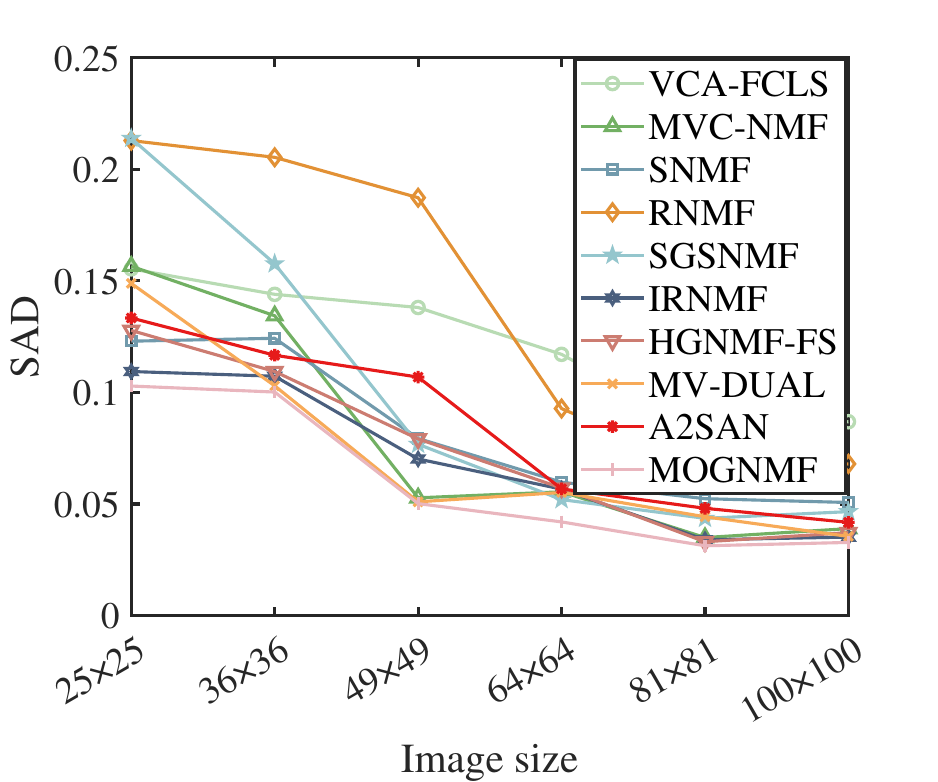}
	}
	\subfigure[ ]{			
		\includegraphics[scale=0.26]{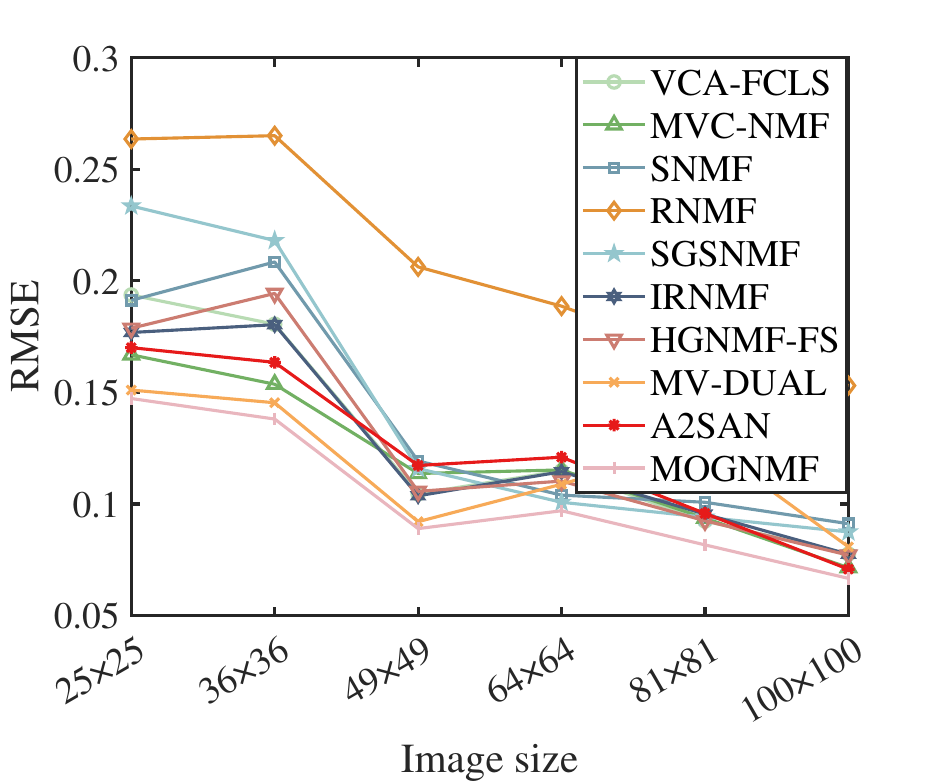}
	}
	\vspace{-0.4cm}
	\caption{Impact of image sizes, where (a) SAD and (b) RMSE.}\label{imagesize}
\end{figure}

This experiment aims to evaluate the sensitivity to different image sizes, where the SNR is set to 20 dB, the number of endmembers is fixed to 9, and the image sizes are set to  \(25 \times 25\), \(36 \times 36\), \(49 \times 49\), \(64 \times 64\), \(81 \times 81\), and \(100 \times 100\). As shown in Fig. \ref{imagesize}, as the number of pixels increases, SGSNMF achieves better unmixing performance than SNMF, which is mainly attributed to its integration of spatial group structure and abundance sparsity. Our proposed MOGNMF, by incorporating the multi-order graph and dual sparsity, demonstrates greater adaptability and unmixing capability when handling HSIs with different pixel counts.
\begin{figure}[t]
	\setlength{\subfigcapskip}{-2pt}  
	\subfigure[ ]{
		\includegraphics[scale=0.26]{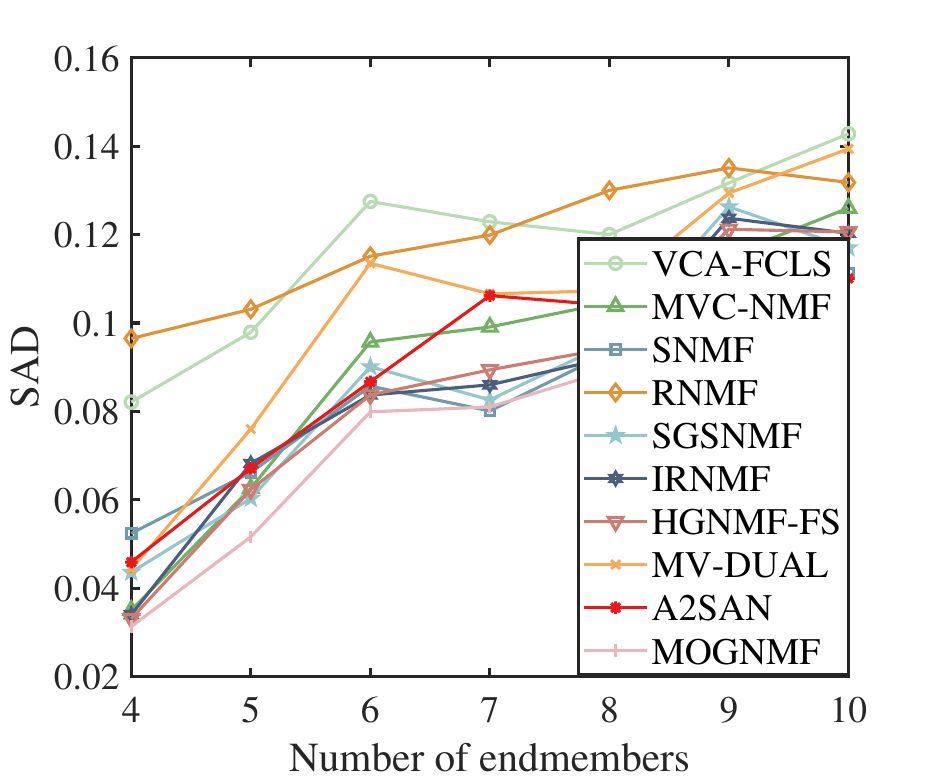}
	}
	\subfigure[ ]{			
		\includegraphics[scale=0.26]{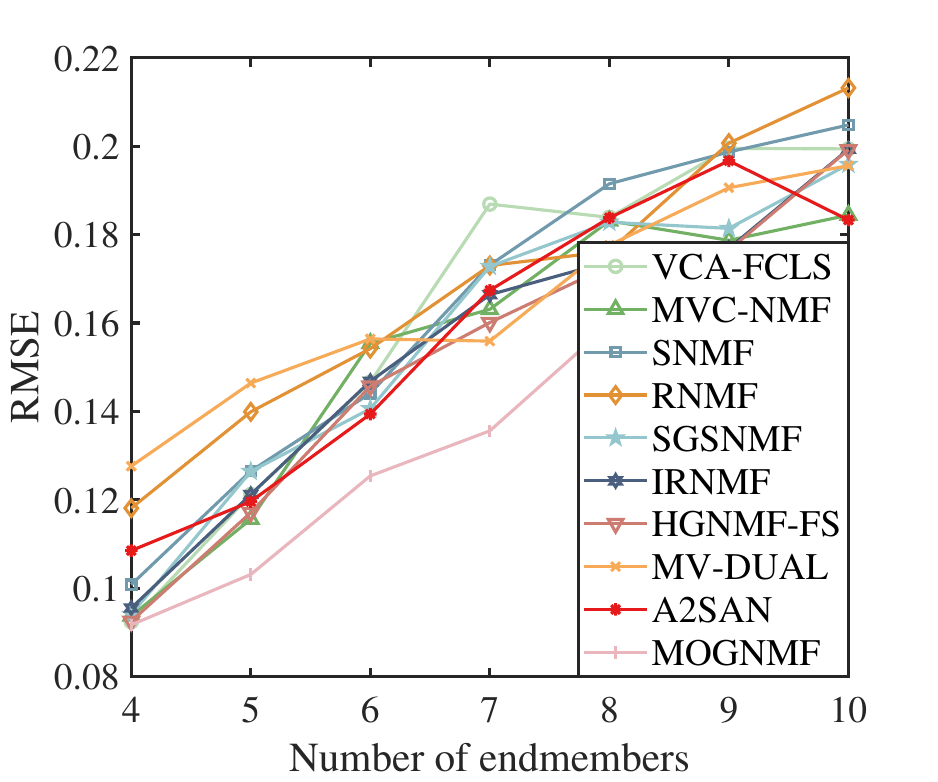}
	}
	\vspace{-0.4cm}
	\caption{Impact of endmember numbers, where (a) SAD and  (b) RMSE.}\label{duanyuanshu}
\end{figure}

\begin{figure}[t]
	\subfigcapskip=-5pt
	\subfigure[ ]{
		\includegraphics[scale=0.26]{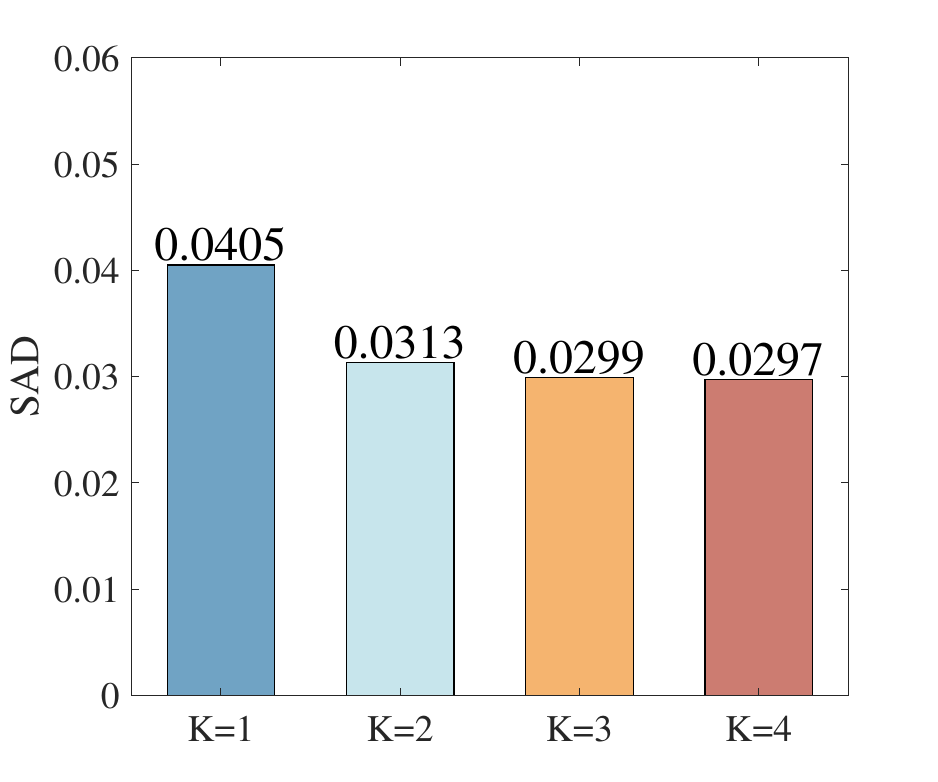} 
	}\subfigcapskip=-5pt
	\subfigure[ ]{			
		\includegraphics[scale=0.26]{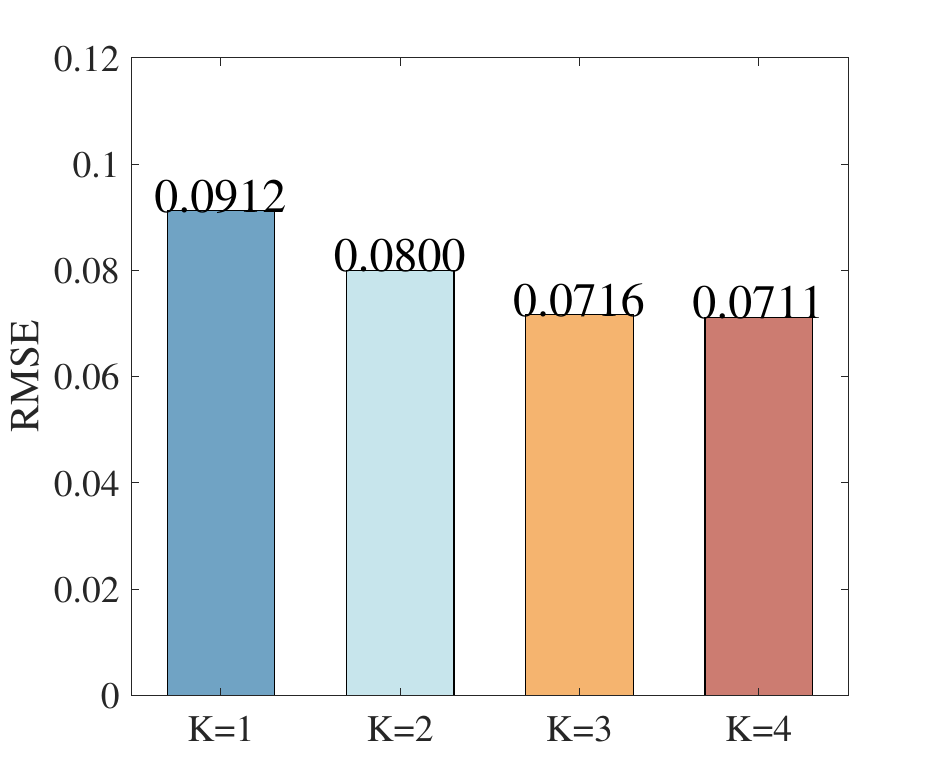}
	}	\subfigcapskip=-5pt
	\vspace{-0.2cm}
	\caption{Impact of order $K$ on the Simu-1 dataset, where (a) SAD and (b) RMSE.}\label{fig:K}
\end{figure}

\begin{figure}[t]
	\setlength{\subfigcapskip}{-5pt}  
	\subfigure[ ]{
	\includegraphics[scale=0.26]{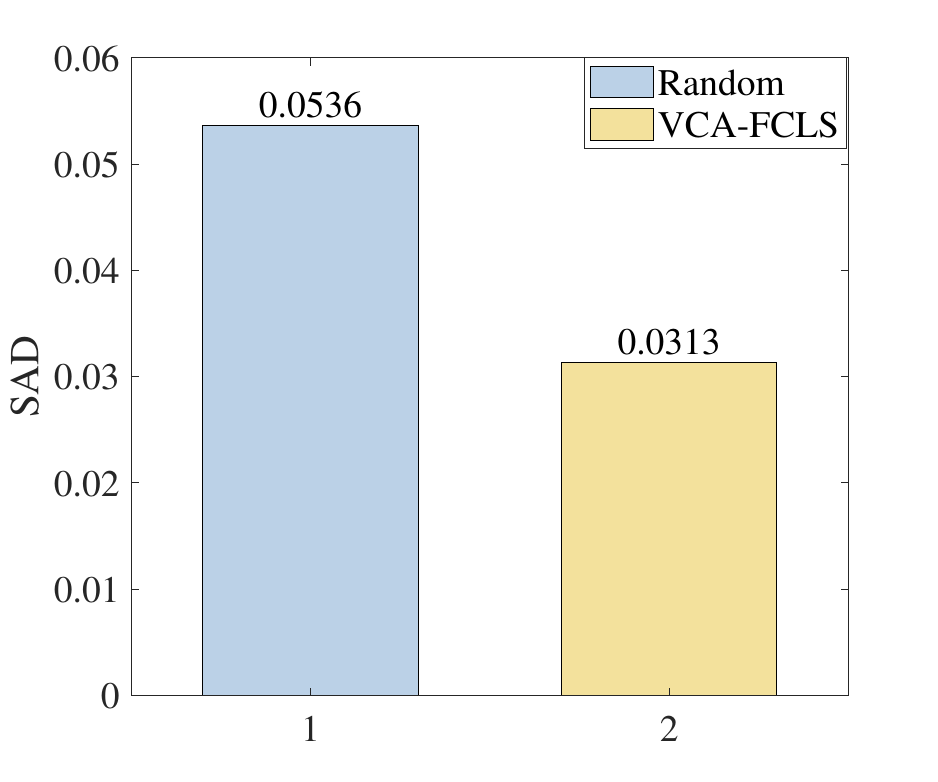} 
	}
	\subfigure[ ]{			
		\includegraphics[scale=0.26]{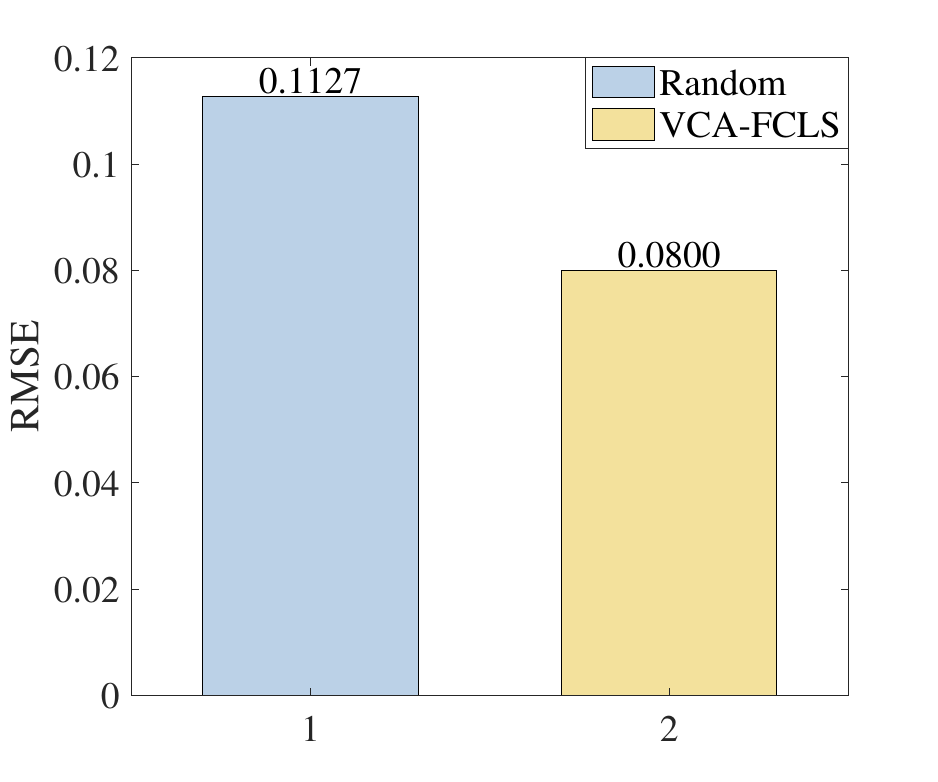} 
	}
	\vspace{-0.4cm}
	\caption{Comparison with different initializations, where (a) SAD and (b) RMSE.}\label{Sensitivity} 
\end{figure}

\begin{figure}[t]
    \setlength{\subfigcapskip}{-4pt} 
    \centering
    \subfigure[ ]{
        \includegraphics[scale=0.259]{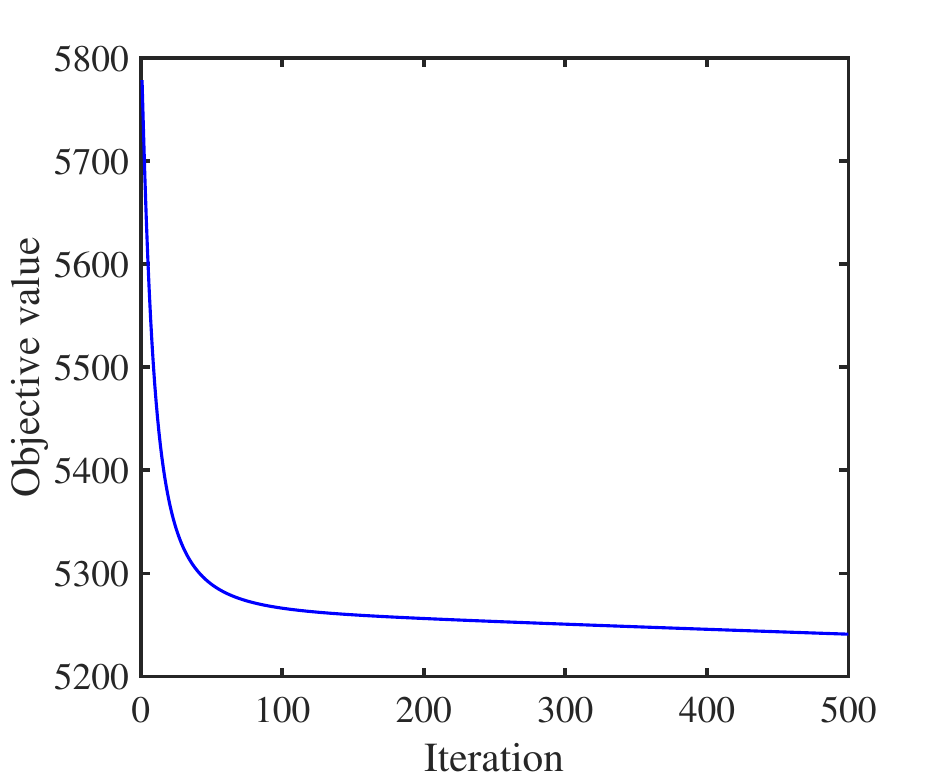} 
    }
    \subfigure[ ]{
        \includegraphics[scale=0.259]{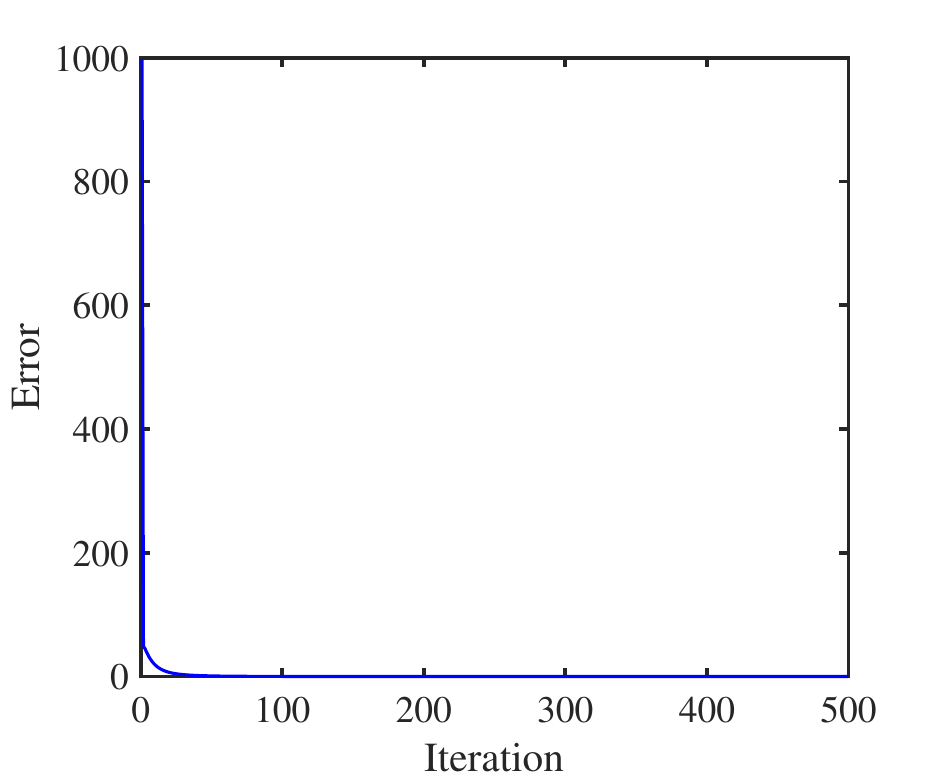} 
    }
    \vspace{-0.4cm}
	\caption{Convergence analysis, where (a) Objective values and (b) Error.}\label{Convergence} 
\end{figure}

To evaluate the performance under different  numbers of endmembers, the SNR is set to 20 dB, and the number of endmembers ranges from 4 to 10. As shown in Fig. \ref{duanyuanshu}, the overall unmixing accuracy improves when the number of endmembers increases. This is because a larger number of endmembers introduces greater spectral mixing complexity, requiring the separation of more spectral features. Nevertheless, our proposed MOGNMF outperforms other compared methods in most cases, illustrating greater potential in unmixing.
     
\subsubsection{Impact of Order $K$} 
To analyze the influence of graph order \( K \) on unmixing performance, Algorithm~\ref{getgraph} is evaluated with \( K = 1, 2, 3 \). Specifically, \( K=1 \) corresponds to the use of first-order spectral and spatial graphs, \( K=2 \) includes both first- and second-order graphs, and so on, forming a multi-order graph structure that captures increasingly complex pixel relationships. As shown in Fig.~\ref{fig:K}, unmixing results improve as the graph order \( K \) increases. This improvement can be attributed to the fact that a first-order graph only captures local neighborhood information. When \( K = 2 \), the graph incorporates neighbors of neighbors, and with \( K = 3 \), the information flow within the graph is further expanded across more distant nodes. The use of higher-order graphs thus allows the model to integrate more comprehensive spatial–spectral information and introduces beneficial cross-region redundancy. This extended connectivity contributes to smoother abundance estimations by encouraging consistency among spectrally similar but spatially distant pixels. However, although increasing the graph order \( K \) from 1 to 3 leads to significant performance improvements, further increasing the order, such as to \( K = 4 \), brings only marginal gains. This suggests that the structural patterns of higher-order graphs gradually become similar to those of their lower-order counterparts, offering limited additional information. In other words, higher-order graphs beyond the third order do not contribute substantially new or beneficial structural features, as the third-order graph is already sufficient to capture the underlying data structure effectively.

\subsubsection{Sensitivity Analysis}
To investigate the impact of different initialization strategies on the  unmixing performance, random initialization and VCA-FCLS initialization are compared. The experiments are conducted on the synthetic dataset with the SNR of 20 dB and six endmembers. As shown in Fig.~\ref{Sensitivity}, the VCA-FCLS  initialization achieves better unmixing performance compared to the random initialization. This is mainly because VCA can exploit the geometric structure of hyperspectral data to extract potential endmembers, while FCLS can estimate the abundance matrix under nonnegativity and sum-to-one constraints, ensuring physical interpretability.

\subsubsection{Robustness Analysis} 
Table  \ref{SAMONTABLE} illustrates the performance of our proposed MOGNMF and other comparison methods on the  Samson dataset under different SNR conditions. Since  VCA-FCLS does not consider noise or spatial information and heavily relies on the pure pixel assumption, it is sensitive to noise and its performance degrades under low SNR conditions. In contrast, IRNMF, which constructs a group sparse constrained NMF, is more robust to noise than RNMF and achieves relatively good unmixing results under various SNR levels. HGNMF-FS leverages multi-order graph modeling to represent relationships among HSI pixels, which improves its noise resistance. As mentioned above, MOGNMF constructs more accurate multi-order graphs, thereby demonstrating stronger resistance to interference. Overall, our proposed method consistently achieves the best or second-best unmixing results across all SNR conditions, indicating its high accuracy in unmixing and strong robustness under different noise levels.

\subsubsection{Convergence Analysis} \label{Convergence Analysis}

This study is conducted with 9 endmembers and the SNR ratio of 20. Fig.  \ref{Convergence} shows that as the number of iterations increases, the objective function value decreases rapidly and gradually stabilizes. In addition, the relative error versus the number of iterations is provided, demonstrating good convergence behavior.

\section{Conclusion}\label{sec:CONCLUSION}

In this paper, we have provided a novel HU method called adaptive multi-order graph regularized NMF (MOGNMF), which successfully introduces multi-order graph regularization from both spatial and spectral perspectives, and obtains the optimal higher-order Laplacian through iterative optimization. This method effectively addresses the challenges of order and parameter selection in traditional graph learning-based NMF methods.  Extensive verification shows  the superior performance of such adaptive multi-order graph  learning. In addition, dual sparsity with $\ell_{1/2}$-norm  and $\ell_{2,1}$-norm is incorporated, and its effectiveness has been verified by ablation experiments. To our best  knowledge, this is the first time that integrates adaptive multi-order graph  and dual sparsity into the NMF framework, with promising performance in HU.

However, introducing high-order graph relationships inevitably increases computational and memory overhead. In the future, we are interested in developing lightweight network architectures to reduce complexity while maintaining accuracy.

	\bibliographystyle{IEEEtran}
	\bibliography{mybibfile}

\end{document}